\PassOptionsToPackage{dvipsnames}{xcolor}
\documentclass{article} 
\usepackage{iclr2027_conference,times}
\usepackage{amsmath,amssymb}

\usepackage{amsmath,amsfonts,bm}

\def\eqref#1{equation~\ref{#1}}

\def\1{\bm{1}}

\DeclareMathAlphabet{\mathsfit}{\encodingdefault}{\sfdefault}{m}{sl}
\SetMathAlphabet{\mathsfit}{bold}{\encodingdefault}{\sfdefault}{bx}{n}

\usepackage{url}

\makeatletter
\newcommand{\blfootnote}[1]{%
  \begingroup
  \renewcommand{\thefootnote}{}%
  \renewcommand{\@makefntext}[1]{\noindent##1}%
  \begin{NoHyper}\footnotetext{#1}\end{NoHyper}%
  \endgroup
}
\makeatother

\usepackage{colortbl}
\usepackage{graphicx}
\usepackage{xspace}
\usepackage[utf8]{inputenc} 
\usepackage[T1]{fontenc}    
\usepackage[dvipsnames]{xcolor}
\usepackage[
  colorlinks=true,
  linkcolor=NavyBlue,
  citecolor=TealBlue,
  urlcolor=RoyalBlue
]{hyperref}
\usepackage{bbm}
\usepackage{enumitem}     
\usepackage{booktabs}     
\usepackage{algorithm}    
\usepackage{algorithmic}  
\usepackage{natbib}       
\usepackage{inconsolata}    
\usepackage{booktabs}
\usepackage{xcolor}
\usepackage{wasysym}
\usepackage{threeparttable}
\usepackage{graphicx}
\usepackage{array}
\usepackage{colortbl}
\usepackage{tikz}
\usepackage{titletoc}
\usepackage{subcaption}
\usepackage[most]{tcolorbox}
\usepackage{enumitem}
\usepackage{changepage}
\usepackage{caption}
\usepackage{fvextra}
\usepackage{tabularx}
\usepackage{needspace}   
\tcbuselibrary{skins}

\definecolor{PromptBackground}{RGB}{248,249,251}
\definecolor{PromptFrame}{RGB}{200,208,220}
\definecolor{PromptSystem}{RGB}{37,82,140}
\definecolor{PromptUser}{RGB}{34,120,78}
\definecolor{PromptVariable}{RGB}{190,60,30}

\newcommand{\PromptTextSetup}{%
  \setlength{\parindent}{0pt}%
  \setlength{\parskip}{0.3em}%
  \setlist[itemize]{leftmargin=1.4em,itemsep=0.2em,topsep=0.2em,
    parsep=0pt,partopsep=0pt}%
  \setlist[enumerate,1]{leftmargin=1.9em,labelwidth=1.2em,labelsep=0.4em,
    align=left,itemsep=0.2em,topsep=0.2em,parsep=0pt,partopsep=0pt}%
}

\tcbset{
  paper prompt/.style={
    enhanced,breakable,
    width=\linewidth,
    colback=PromptBackground,colframe=PromptFrame,
    boxrule=0.6pt,arc=4pt,
    boxsep=0pt,left=10pt,right=10pt,top=12pt,bottom=8pt,
    fontupper=\footnotesize\sffamily\raggedright,
    fonttitle=\footnotesize\sffamily\bfseries,
    coltitle=white,
    titlerule=0pt,toptitle=3pt,bottomtitle=3pt,
    lefttitle=10pt,righttitle=10pt,
    attach boxed title to top left={xshift=10pt,yshift=-7pt},
    boxed title style={
      boxrule=0pt,arc=2pt,boxsep=0pt,
      left=6pt,right=6pt,top=3pt,bottom=3pt
    },
    before skip=9pt,after skip=4pt,
    pad at break*=6pt,
    lines before break=4,
    before upper=\PromptTextSetup
  },
  system prompt/.style={
    paper prompt,colbacktitle=PromptSystem,
    before upper={\PromptTextSetup\colorlet{PromptAccent}{PromptSystem}},
    title={System Prompt: #1},
    title after break={System Prompt: #1 (continued)}
  },
  user prompt/.style={
    paper prompt,colbacktitle=PromptUser,
    before upper={\PromptTextSetup\colorlet{PromptAccent}{PromptUser}},
    title={User Prompt: #1},
    title after break={User Prompt: #1 (continued)}
  },
  prompt excerpt/.style={
    enhanced,colback=white,colframe=white,boxrule=0pt,arc=0pt,
    borderline west={1.5pt}{0pt}{PromptUser!45!PromptFrame},
    boxsep=0pt,left=7pt,right=5pt,top=5pt,bottom=5pt,
    before skip=4pt,after skip=4pt,
    fontupper=\footnotesize\sffamily\raggedright
  }
}

\newcommand{\PromptHeading}[1]{%
  \par\addvspace{0.6\baselineskip}%
  {\bfseries\color{PromptAccent}#1\par}%
  \nobreak\smallskip\nobreak
}
\newcommand{\PromptSubheading}[1]{%
  \par\addvspace{0.4\baselineskip}%
  {\bfseries #1\par}%
  \nobreak\smallskip\nobreak
}

\DefineVerbatimEnvironment{PromptCode}{Verbatim}{
  fontsize=\scriptsize,
  samepage=true,
  breaklines=true,breakanywhere=false,
  breaksymbolleft={},breaksymbolright={},
  frame=leftline,rulecolor=\color{PromptFrame},
  framerule=1pt,framesep=6pt
}

\title{Forecast-Dojo: Replayable Environments for Benchmarking and Training LLM Forecasting Agents}

\newcommand{\projectdataurl}{https://huggingface.co/datasets/liqinye/Forecast-Dojo}
\newcommand{\projectcodeurl}{https://github.com/Forecast-Dojo/Forecast-Dojo}
\newcommand{\projectwebsiteurl}{https://forecast-dojo.github.io/}

\definecolor{ProjectLink}{HTML}{234F9F}
\newcommand{\projectresource}[3]{%
    \begingroup
    \edef\resourceurl{#1}%
    \def\resourcelabel{%
        \mbox{%
            \makebox[1.45em][c]{%
                \raisebox{-0.2\height}{%
                    \includegraphics[height=1.25em]{figures-arxiv/icons/#2.pdf}%
                }%
            }%
            \hspace{0.35em}\textcolor{ProjectLink}{\texttt{#3}}%
        }%
    }%
    \ifx\resourceurl\empty
        \resourcelabel
    \else
        \href{\resourceurl}{\resourcelabel}%
    \fi
    \endgroup
}
\newcommand{\projectresources}{%
    \raisebox{-5pt}[\height][\depth]{%
    \normalfont\sffamily\fontsize{9}{11}\selectfont
    \projectresource{\projectdataurl}{huggingface}{Data}%
    \hspace{1.5em}%
    \projectresource{\projectcodeurl}{github}{Code}%
    \hspace{1.5em}%
    \projectresource{\projectwebsiteurl}{website}{Website}}%
}

\author{%
 Liqin Ye\textsuperscript{*,1}, 
 Haorui Wang\textsuperscript{*,1}, 
 Fardin Ahmed\textsuperscript{1},
 \textbf{Rongzhi Zhang}\textsuperscript{\textdagger,2}, 
 \textbf{Yuan He}\textsuperscript{\textdagger,2}, 
 \textbf{Ziyuan Lin}\textsuperscript{3}, \\
 \textbf{Yanbin Yin}\textsuperscript{1},
 \textbf{Jing Peng}\textsuperscript{1}, 
 \textbf{Michael Galarnyk}\textsuperscript{1},
 \textbf{Sudheer Chava}\textsuperscript{1}, 
 \textbf{Chao Zhang}\textsuperscript{1}\\[3pt]
 \textsuperscript{1}Georgia Institute of Technology,
 \textsuperscript{2}Amazon, \textsuperscript{3}University of Florida\\[7pt]
  \makebox[\dimexpr\textwidth-2\tabcolsep\relax][l]{\projectresources}
}

\newcommand{\ours}{\texttt{Forecast-Dojo}\xspace}

\iclrfinalcopy 
\begin{document}

\maketitle
\lhead{Preprint}  
\blfootnote{\textsuperscript{*}Equal contribution. Corresponding author(s): \texttt{\{liqiny, hwang984\}@gatech.edu}}
\blfootnote{\textsuperscript{\textdagger}Work done outside Amazon.}
\begin{abstract}
Forecasting requires large language model (LLM) agents to gather evolving evidence, revise their beliefs, and make calibrated predictions over time. Yet evaluating and improving these capabilities is challenging because real-world events unfold continuously and outcomes may take months to resolve. We introduce \ours, a replayable environment that reconstructs resolved prediction-market events as sequences of dated forecasting steps, enabling reproducible evaluation, longitudinal analysis, and training from agent interactions. It comprises 1,568 Polymarket events split temporally into training and evaluation sets, alongside 18.8 million dated news articles. Across 12 models, research tools consistently improve forecasting performance, and greater capture of new evidence is associated with larger improvements over time. Nevertheless, all models trail historical market forecasts, while carrying a belief notebook across dates reduces research costs without consistently improving forecast quality. We supervise fine-tuned Qwen3-30B-A3B-Thinking-2507 on collected trajectories from Qwen3-235B-A22B-Thinking-2507 to demonstrate the environment's training utility. The student's accuracy is improved from 34.8\% to 42.8\% on the evaluation set.
\end{abstract}

\begin{figure}[!ht]
  \centering
  \includegraphics[width=\linewidth]{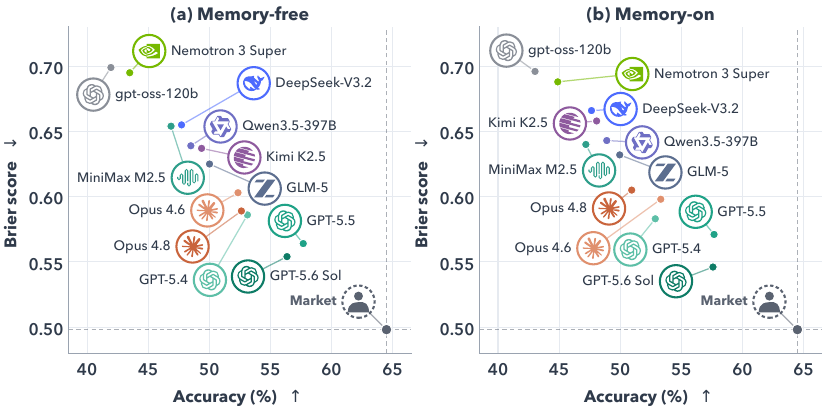}%
  \vspace{-0.38em}
  \caption{\textbf{Forecasting performance on \ours.}
Accuracy and Brier score of 12 models on 230 held-out events.
Both settings use research tools: (a) \emph{memory-free} starts with a fresh context at each forecast date, while (b) \emph{memory-on} carries a belief notebook across dates. Lower right is better; dashed lines mark historical market performance. Full results are reported in Table~\ref{tab:dojo_main_results}.}
    \label{fig:teaser}
    \vspace{-1.0em}
  \end{figure}

\section{Introduction}
\label{sec:introduction}
Large language model (LLM) agents are increasingly used to forecast real-world events by actively searching for evidence, reasoning under uncertainty, and producing probabilistic predictions~\citep{approachHuman,futurex,futurexpro,forecastbench}. Forecasting is inherently time-dependent: the evidence available to a forecaster changes as new information arrives. The same question can therefore pose a different prediction problem at different times, requiring the forecaster to update its belief as new evidence emerges.

Existing forecasting benchmarks capture only part of this process (Table~\ref{tab:related-work}). Live benchmarks~\citep{forecastbench,futurex,livemarcoeval,predictionarena} pose unresolved questions, so the forecasting problem evolves naturally with the world, but they run on wall-clock time. A past forecasting condition cannot be recreated for a model released later, and outcomes arrive only at resolution, which slows evaluation and makes training impractical. Historical benchmarks~\citep{forecastqa,approachHuman,chandak2025scaling} reconstruct past information cutoffs, making resolved events immediately scorable and reusable. However, they typically evaluate each forecast at a single historical point rather than revisiting the same question across multiple points in time. The ideal setting combines the two: the forecasting problem evolves as new evidence becomes available, yet each past step can be replayed with the same task and information cutoff. Models can then be compared under identical conditions and scored immediately, and their forecasting trajectories become usable for training.

\begin{figure}[!ht]
    \centering
    \includegraphics[width=\linewidth]{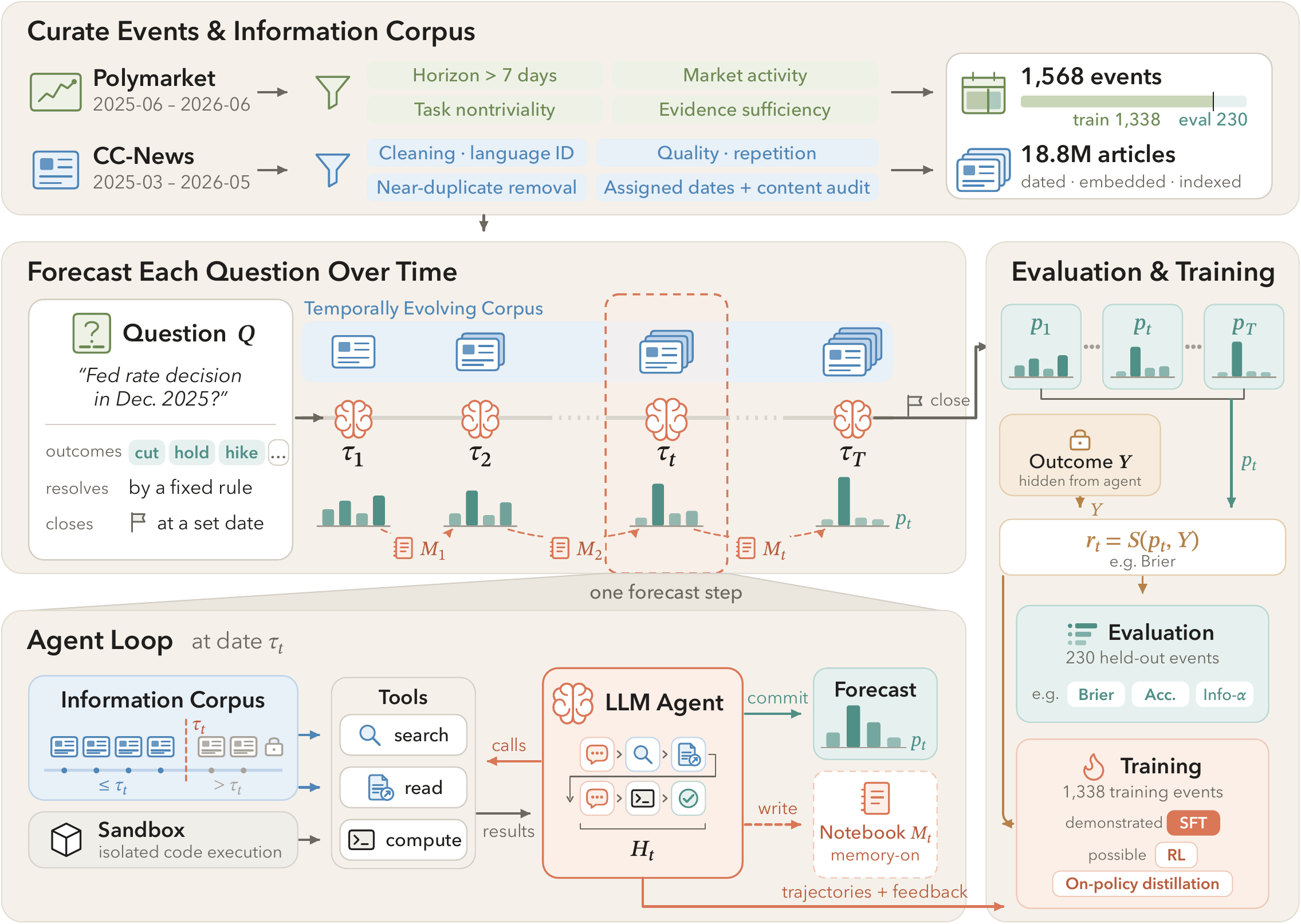}
    \vspace{-1.5em}
    \caption{\textbf{Overview of Forecast-Dojo.}
Top: resolved Polymarket events are filtered and split by time into training and evaluation events, and CC-News articles are cleaned, dated, and indexed.
Middle left: each question is forecast at a fixed sequence of dates as the visible corpus grows, with an optional belief notebook $M_t$ passed between steps.
Bottom left: within one step, the agent searches and reads articles dated on or before $\tau_t$, runs code in a sandbox, and commits a forecast $p_t$.
Right: the realized outcome $Y$ stays hidden from the agent. The evaluator scores each forecast against it, using held-out events for evaluation and training events for learning.}
    \label{fig:forecastdojo}
    \vspace{-1.5em}
\end{figure}

We introduce \ours, a replayable environment that reconstructs resolved real-world events as sequences of historical forecast steps (Figure~\ref{fig:forecastdojo}). At each step, an LLM agent can search a temporally restricted information corpus, inspect full documents, and use computational tools to produce a probabilistic forecast using documents whose assigned dates satisfy that cutoff. Each step can be reset and rerun across models or repeated trials, while steps from the same event can be traversed sequentially with persistent agent memory. The recorded outcome is retained for immediate scoring and is not supplied in the forecasting prompt. This common interaction interface supports both benchmarking and learning: held-out events evaluate agents under controlled conditions, while training events generate forecasting trajectories and outcome feedback for learning.

Empirically, we evaluate 12 models on 230 held-out events without tools, with research tools, and with research tools plus a belief notebook carried between dates. Research tools lower Brier for all 12 models, and memory-free Brier falls from 0.670 in the first third of an event to 0.606 in the last third, while forecasts without tools stay flat. The improvements concentrate at steps with more new evidence. Every model still trails the historical market forecasts in Brier and accuracy, with the best Brier at 0.546 against 0.498 for the market. The belief notebook reduces estimated API costs by a median of 24\% across five proprietary models with available pricing
(Section~\ref{sec:experiments:cost_analysis}). However, it lowers Brier scores for only 6 of the 12 evaluated models, with none of these improvements statistically significant (Section~\ref{sec:experiments:performance}). Finally, a supervised fine-tuning study demonstrates that Qwen3-30B-A3B-Thinking-2507~\citep{yang2025qwen3}, trained on trajectories collected from Qwen3-235B-A22B-Thinking-2507~\citep{yang2025qwen3} in the \ours training set, outperforms its base model on temporally held-out events (Section~\ref{sec:experiments:training}).

Our contributions are:
\begin{itemize}[leftmargin=1.2em, itemsep=0.1em, topsep=-0.2em]
    \item \textbf{A replayable forecasting environment.} \ours replays each resolved event as a fixed sequence of dated forecast steps with an optional belief notebook, and scores every step against the realized outcome for evaluation or training.
    \item \textbf{A dataset of resolved events and dated news.} It covers 1,568 Polymarket events with 6,122 forecast steps, split by time into 1,338 training and 230 evaluation events, with evidence from 18.8M dated CC-News articles.
    \item \textbf{A benchmark study beyond overall rankings.} We compare 12 models with historical market forecasts, examine how forecasts change as events unfold and how memory affects quality and cost, and show how the collected interactions can be used to train an agent.
\end{itemize}
\vspace{-0.5em}
\section{Related Work}
\label{sec:related_work}
\vspace{-0.5em}
\paragraph{Forecasting with dated evidence.}
ForecastQA restricts news by time, while Autocast pairs forecasting questions with dated articles and historical human forecasts \citep{forecastqa,zou2022autocast}.
ForecastBench and FutureX collect predictions on unresolved events \citep{forecastbench,futurex}.
Related settings evaluate repeated forecasts, macroeconomic nowcasts, and simulated market decisions \citep{yang2025prophetarena,livemarcoeval,cheng2026polybench}.
\ours uses resolved events to support repeatable research and immediate outcome-based feedback, while keeping market probabilities outside the forecasting prompt.

\begin{table}[t]
\centering
\begingroup
\small
\setlength{\tabcolsep}{2pt}
\renewcommand{\arraystretch}{1.15}

\definecolor{dojotint}{HTML}{EAF2FA}
\definecolor{supportgreen}{HTML}{4E8A83}
\definecolor{unsupportedred}{HTML}{B87575}
\definecolor{supportgray}{HTML}{7C8792}

\DeclareRobustCommand{\hasfeature}{%
  \tikz[baseline=.05em,x=1em,y=1em,line width=.2em,
        line cap=round,line join=round]{%
    \path[use as bounding box] (0,0) rectangle (.95,.75);
    \draw[supportgreen] (.12,.35) -- (.37,.12) -- (.84,.65);}}

\DeclareRobustCommand{\notreported}{%
  \tikz[baseline=.05em,x=1em,y=1em,line width=.2em,
        line cap=round]{%
    \path[use as bounding box] (0,0) rectangle (.95,.75);
    \draw[unsupportedred] (.23,.16) -- (.73,.66)
                         (.23,.66) -- (.73,.16);}}

\DeclareRobustCommand{\customsystem}{%
  \tikz[baseline=.05em,x=1em,y=1em,line width=.2em,
        line cap=round,line join=round]{%
    \path[use as bounding box] (0,0) rectangle (.95,.75);
    \draw[supportgray] (.12,.35) -- (.37,.12) -- (.84,.65)
                      (.47,.535) -- (.755,.25);}}

\newcolumntype{C}{>{\centering\arraybackslash}m{\dimexpr(.45\linewidth-14\tabcolsep)/4\relax}}

\caption{\textbf{Summary of existing forecasting benchmarks.}
The criteria target the capabilities used to construct and study Forecast-Dojo.
\emph{Aligned evaluation} indicates that the same forecasting task is evaluated
at a pre-specified sequence of forecast dates or evidence states shared across models.
\emph{Market belief} indicates that historical prediction-market prices are available as a reference.
\hasfeature: provided; \notreported: not documented in the cited work; \customsystem: system- or agent-dependent.}
\vspace{-0.8em}
\label{tab:related-work}

\begin{tabular}{
>{\raggedright\arraybackslash}m{.34\linewidth}
CC
>{\centering\arraybackslash}m{.11\linewidth}
CC
>{\centering\arraybackslash}m{.10\linewidth}
}
\toprule
\textbf{Benchmark}
& \shortstack{\textbf{Dated}\\\textbf{replay}}
& \shortstack{\textbf{Agent}\\\textbf{research}}
& \shortstack{\textbf{Aligned}\\\textbf{evaluation}}
& \shortstack{\textbf{Agent}\\\textbf{memory}}
& \shortstack{\textbf{Market}\\\textbf{belief}}
& \shortstack{\textbf{Train/eval}\\\textbf{split}} \\
\midrule

ForecastQA~\citep{forecastqa}
& \hasfeature
& \notreported
& \notreported
& \notreported
& \notreported
& \hasfeature \\

Autocast~\citep{zou2022autocast}
& \hasfeature
& \notreported
& \hasfeature
& \notreported
& \notreported
& \hasfeature \\

MIRAI~\citep{ye2024mirai}
& \hasfeature
& \hasfeature
& \hasfeature
& \notreported
& \notreported
& \notreported \\

ForecastBench~\citep{forecastbench}
& \notreported
& \customsystem
& \notreported
& \notreported
& \notreported
& \notreported \\

FutureX~\citep{futurex}
& \notreported
& \hasfeature
& \notreported
& \notreported
& \notreported
& \notreported \\

Prophet Arena~\citep{yang2025prophetarena}
& \notreported
& \notreported
& \hasfeature
& \notreported
& \hasfeature
& \notreported \\

EvolveCast~\citep{yuan2025language}
& \hasfeature
& \notreported
& \hasfeature
& \notreported
& \notreported
& \notreported \\

BTF-2~\citep{liptay2026evaluating}
& \hasfeature
& \hasfeature
& \notreported
& \notreported
& \notreported
& \notreported \\

FutureSim~\citep{goel2026futuresim}
& \hasfeature
& \hasfeature
& \customsystem
& \hasfeature
& \notreported
& \notreported \\

\rowcolor{dojotint}
\textbf{\ours}
& \hasfeature
& \hasfeature
& \hasfeature
& \hasfeature
& \hasfeature
& \hasfeature \\

\bottomrule
\end{tabular}
\endgroup
\end{table}

\vspace{-0.75em}
\paragraph{Interactive research and forecast revision.}
Bench to the Future uses frozen research corpora, and BTF-2 records traces to distinguish information gathering from judgment \citep{wildman2025bench,liptay2026evaluating}.
MIRAI provides code-based access to events and news, while WorldReasoner evaluates outcome, evidence, and reasoning quality \citep{ye2024mirai,chi2026worldreasoner}. FutureSim studies long-horizon adaptation in a replayed world, where agents jointly decide how to research, maintain memory, revise forecasts, and progress through time \citep{goel2026futuresim}. \ours instead fixes time progression in the experimental design: each event is replayed at the same historical checkpoints for all agents, which yields matched longitudinal trajectories for controlled comparison and reusable interactions with outcome feedback for learning. This complements work on evidence-driven revision, probability coherence, and iterative research workflows \citep{yuan2025language,paleka2025consistency,murphy2026agentic}.

\vspace{-0.75em}
\paragraph{Environments for training agents.}
MLE-Dojo provides executable machine learning engineering tasks and feedback for evaluation and training \citep{qiang2025mledojo}.
\ours follows this environment-centered approach, specifying the forecasting interaction and its feedback while leaving learning algorithms separate.
Prior forecasting work trains on question collections or selected model-generated forecasts \citep{zou2022autocast,approachHuman}, and recent methods use reinforcement learning, outcome-based fine-tuning, and news-derived questions \citep{levy2026reinforcement,jeen2026reaching,chandak2025scaling}.
Our contribution is the shared task and tool interface for collecting research interactions and evaluating agents on separate events; the SFT experiment demonstrates one use of that interface.

\section{Forecasting as an Interactive Task}
\label{sec:task}

We formulate forecasting as an interactive task over a sequence of forecast steps for the same event. At each step, the agent researches evidence available up to the current date and produces a probabilistic forecast. Across steps, the evidence boundary advances and explicit memory may persist. 

\paragraph{Events, forecast steps, and episodes.}
A question $Q$ specifies an event, its resolution criteria, and a finite set of mutually exclusive outcomes $\mathcal Y=\{1,\ldots,K\}$. Let $Y\in\mathcal Y$ denote the realized outcome. At ordered dates $\tau_1<\cdots<\tau_T$ before the event closes, where $T$ varies across events, the agent performs a \emph{forecast step}: it researches the event and reports a probability distribution $p_{Q,t}\in\Delta^{K-1}$. The ordered forecast steps for the same question form a \emph{forecast episode}. All steps concern the same eventual outcome, but they differ in the historical evidence available at the forecast date. 

\paragraph{Within-step interaction.}
Let $\mathcal I$ denote an information corpus. At date $\tau_t$, the agent is exposed to
\begin{equation}
    \mathcal I_{\leq\tau_t}
    =\{d\in\mathcal I:\operatorname{date}(d)\leq\tau_t\},
    \qquad
    \mathcal I_{\leq\tau_t}\subseteq\mathcal I_{\leq\tau_{t+1}},
    \label{eq:dojo_dated_corpus}
\end{equation}
where $\operatorname{date}(d)$ denotes the timestamp assigned to document $d$. Within a forecast step, the agent may issue search queries, inspect retrieved articles, and use computation before submitting its forecast (see Section~\ref{sec:method:interaction}). Let \(H_t\) denote the within-step interaction history, including the agent's research actions and the resulting tool observations. The agent produces \(p_{Q,t}\) conditioned on \(H_t\). Because agents choose their own queries and which documents to inspect, different rollouts at the same date from an agent can follow different research paths even under the same dated evidence boundary.

\paragraph{Progression across forecast steps.}
After each report, the environment advances to the next scheduled date and starts a fresh interaction. We define two modes for carrying information across forecast steps: \emph{memory-free} and \emph{memory-on}. In memory-free forecasting, each step starts without information produced at previous steps. In memory-on forecasting, the agent produces a belief notebook $M_t$ that summarizes its current assessment, supporting evidence, and open questions. At the next step, $M_t$ is provided alongside the question and new forecast date, allowing the agent to update its forecast from its prior assessment as new evidence becomes available.

\paragraph{Outcome feedback.}
Each forecast step ends with a probability report $p_{Q,t}$. Once the realized outcome $Y$ is available, the environment assigns feedback $r_{Q,t}=S(p_{Q,t},Y)$,
where $S$ is an outcome-based scoring rule. The outcome and feedback are not part of the agent's forecasting context. This separates the interaction that produces a forecast from the feedback assigned to it, allowing the same task interface to support different downstream evaluation or learning procedures.
\section{Forecast-Dojo}
\label{sec:env}

Section~\ref{sec:task} defines the forecasting task abstractly. We now describe how events (\S\ref{sec:method:events}), information corpus (\S\ref{sec:method:corpus}), and forecast dates (\S\ref{sec:method:episodes}) are selected, how agents interact with research tools and optionally carry memory across steps (\S\ref{sec:method:interaction}), and how the resulting forecasts and trajectories support evaluation and learning (\S\ref{sec:method:evaluation}).

\subsection{Forecasting Events}
\label{sec:method:events}

We construct our candidate pool from resolved Polymarket\footnote[1]{\url{https://polymarket.com/}} binary and mutually exclusive multi-option events whose full lifetimes fall between June 2025 and June 2026. We construct a daily market reference from UTC-day mean YES prices, which we call the \emph{market belief}. For binary events, the YES price determines the probability of YES and its complement; for multi-option events, we normalize the option-level YES prices to obtain a distribution over the mutually exclusive outcomes.

Within this pool, we select events along four dimensions. (1) \emph{Forecasting horizon}: the tradeable lifetime must exceed 7 days, so that each question supports multiple forecast steps. (2) \emph{Market activity}: most days must have sufficient trading activity, which reduces sparse or stale prices and favors questions with sustained attention. (3) \emph{Task nontriviality}: we remove events whose market already assigns near-certain probability to the realized outcome, as well as events whose market history is nearly flat and directionless. (4) \emph{Evidence sufficiency}: following \citet{joren2024sufficient}, an LLM judge must rate the evidence retrieved two days before the recorded close as \textsc{Sufficient}, followed by a content audit. This criterion does not establish sufficient evidence at earlier forecast dates. Several filters use the realized outcome, so the benchmark is not a random sample of Polymarket events (Appendices~\ref{sec:appendix:data:event} and~\ref{sec:appendix:selection_sensitivity}).

\subsection{Information Corpus}
\label{sec:method:corpus}

\vspace{-0.3em}
\paragraph{Corpus construction.}
\ours requires broad historical evidence whose availability can be reconstructed at each forecast date. We build the corpus from CC-News~\citep{nagel2016ccnews}, which provides large-scale news coverage together with crawl timestamps that support historical reconstruction. We process CC-News archives from March 2025 through May 2026 using a quality pipeline adapted from FineWeb~\citep{fineweb}, including text cleaning, language identification, repetition and document-quality filtering, and near-duplicate removal. We suppress duplicate URLs, identical titles within a seven-day window, and near-verbatim body matches to reduce repeated coverage. After filtering, the corpus contains approximately 18.8M articles, which we embed with Qwen3-Embedding-8B~\citep{qwen3embedding} and index with FAISS~\citep{faiss}.

\vspace{-0.3em}
\paragraph{Temporal integrity.}
We assign each article a UTC calendar day $t(d)$ using the closest-to-crawl rule. Among valid structured publication and modification timestamps, we choose the one closest to the crawl time (breaking ties toward the later timestamp), cap it at the crawl time, and use its UTC day. If neither structured timestamp is valid, we use the crawl day. These timestamps come from standard structured metadata fields~\citep{guha2016schema,schemaDatePublished,googleArticleStructuredData}. Search and article reads admit $t(d)\leq\tau_t$, so articles assigned to the forecast day itself are included. This is a day-level metadata boundary; it does not certify intraday availability or reconstruct earlier versions of an edited page. Event construction additionally screens for content-level outcome leakage. Appendix~\ref{sec:appendix:data:corpus} gives the full dating rule.

\subsection{Temporal Task Construction}
\label{sec:method:episodes}

\vspace{-0.3em}
\paragraph{Forecast-date selection.}
A forecast step for every day of an event's lifetime would let long-lived events dominate the samples. We therefore use a sublinear schedule: after a two-day buffer before the recorded close, an event with $n$ candidate days receives $T=\operatorname{clamp}\left(\operatorname{round}(\sqrt{n}),3,10\right)$ forecast steps. We partition the event history into $T$ temporal bins and select one date from each, favoring (1) \emph{market-belief movement}, larger changes in the market belief as a proxy for changes in expectations, and (2) \emph{news volume}, more articles in the corpus overall, a global signal rather than an event-specific measure of relevance.

\begin{figure}[t]
    \centering
    \includegraphics[width=\linewidth]{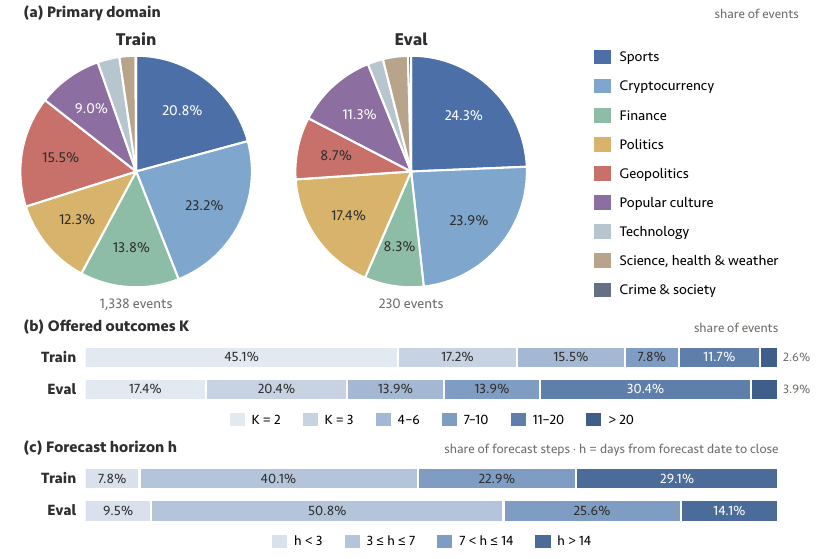}
    \vspace{-1.8em}
    \caption{\textbf{Composition of the training and evaluation splits.}
(a) Primary domain of each event; slices under 5\% are unlabeled.
(b) Number of offered outcomes $K$ per event; binary events have $K=2$.
(c) Forecast horizon $h$, the number of days from a forecast date to the recorded close, over all forecast steps (event--date pairs; 5,325 training, 797 evaluation). Forecast dates stop two days before the close, so the $h<3$ bin is exactly $h=2$.}
    \label{fig:dataset}
\end{figure}

\vspace{-0.3em}
\paragraph{Temporal train--evaluation split.}
We split complete events, rather than individual forecast steps, into non-overlapping temporal windows. Training events must start and close within \texttt{[2025-06, 2026-03)} and evaluation events within \texttt{[2026-03, 2026-06)}, and events crossing either boundary are excluded, so no event appears in both splits. The training split contains 1,338 events (602 binary) with 5,325 forecast steps, while the evaluation split contains 230 events (40 binary) with 797 steps. Figure~\ref{fig:dataset} summarizes the composition of both splits, spanning diverse event domains (Figure~\ref{fig:dataset}a), outcome cardinalities (Figure~\ref{fig:dataset}b), and forecasting horizons (Figure~\ref{fig:dataset}c).

\subsection{Agent Interaction}
\label{sec:method:interaction}
\vspace{-0.3em}
\paragraph{Tool interface.}
At each forecast step, the agent can call three tools. \textsc{Search}, via \texttt{search(query, top\_k)}, returns identifiers, titles, dates, retrieval scores, and snippets for the top-$k$ articles in $\mathcal I_{\leq \tau_t}$. \textsc{Read}, via \texttt{scrape(article\_id)}, returns the full text of a retrieved article. \textsc{Compute}, via \texttt{python(code)}, runs model-generated code for numerical analysis, base-rate estimation, or simulation. Python runs in an isolated sandbox with networking disabled and no access to benchmark data such as the task files, corpus, and index (Table~\ref{tab:tools}). The agent may interleave reasoning with repeated tool calls, and each tool output joins the within-step history $H_t$.
\vspace{-0.3em}
\paragraph{Memory transfer.}
Each forecast step is a fresh model interaction with access to $\mathcal I_{\leq \tau_t}$. In memory-free mode, nothing from earlier steps is carried forward. In memory-on mode, the notebook from the previous step is inserted into the next prompt, while previous turns, reasoning traces, and tool observations are discarded, so the notebook is the only transferred information (format in Figure~\ref{fig:memory_on_prompt}).

\begin{table}[!htb]
\centering\footnotesize\setlength{\tabcolsep}{1.5pt}
\caption{\textbf{Main Results.}
Values average recorded forecasts; $\pm$ denotes the SD across four rollout means. Paired event-bootstrap 95\% CIs are discussed in Appendix~\ref{sec:appendix:uncertainty}. Unusable forecasts receive uniform fallback; accuracy is reported in percent, and Information-$\alpha$ requires available market probabilities. \textbf{Bold} and \underline{underlined} indicate the best and second-best models. Superscripts $\dagger$, $\ddagger$, and $\S$ flag configurations with over 5\% unusable forecasts. Uniform and market forecasts serve as references.}
\vspace{-0.5em}
\label{tab:dojo_main_results}
\begingroup

\renewcommand{\arraystretch}{0.96}
\setlength{\aboverulesep}{1.2pt}
\setlength{\belowrulesep}{1.8pt}
\definecolor{FDLightNoTools}{HTML}{7F7D75}
\definecolor{FDLightFree}{HTML}{4E8A83} 
  \definecolor{FDLightMemory}{HTML}{A56A88}
\definecolor{FDLightBand}{HTML}{EFF3F6}
\providecommand{\msd}[2]{%
  \begin{tabular}[t]{@{}r@{}}
    #1\\[-2.1pt]
    {\fontsize{6.2}{6.6}\selectfont
     \textcolor{gray!85}{\textpm\,#2}}
  \end{tabular}}
\begin{tabular*}{\textwidth}{@{\extracolsep{\fill}}lrrrrrrrrr@{}}
\toprule
& \multicolumn{3}{c}{\textcolor{FDLightNoTools}{No tools}} & \multicolumn{3}{c}{\textcolor{FDLightFree}{Tools, memory-free}} & \multicolumn{3}{c}{\textcolor{FDLightMemory}{Tools, memory-on}} \\
\cmidrule(lr){2-4}\cmidrule(lr){5-7}\cmidrule(lr){8-10}
Model & Brier$\downarrow$ & Acc.$\uparrow$ & Info-$\alpha\uparrow$ & Brier$\downarrow$ & Acc.$\uparrow$ & Info-$\alpha\uparrow$ & Brier$\downarrow$ & Acc.$\uparrow$ & Info-$\alpha\uparrow$ \\
\midrule
\rowcolor{FDLightBand}\multicolumn{10}{@{}l}{\textit{Proprietary}}\\
\rowcolor{white}\addlinespace[2pt]
GPT-5.6 Sol & \msd{$\mathbf{0.650}$}{0.004} & \msd{$\mathbf{47.03}$}{0.5} & \msd{$\mathbf{-0.422}$}{0.006} & \msd{$\mathbf{0.554}$}{0.008} & \msd{\underline{$56.36$}}{0.9} & \msd{$\mathbf{-0.138}$}{0.023} & \msd{$\mathbf{0.546}$}{0.004} & \msd{\underline{$57.59$}}{1.3} & \msd{$\mathbf{-0.104}$}{0.017} \\[0.2pt]
GPT-5.5 & \msd{\underline{$0.698$}}{0.001} & \msd{\underline{$43.85$}}{0.6} & \msd{\underline{$-0.603$}}{0.008} & \msd{\underline{$0.564$}}{0.002} & \msd{$\mathbf{57.69}$}{0.7} & \msd{\underline{$-0.186$}}{0.010} & \msd{\underline{$0.571$}}{0.005} & \msd{$\mathbf{57.68}$}{1.0} & \msd{\underline{$-0.198$}}{0.013} \\[0.2pt]
GPT-5.4 & \msd{$0.708$}{0.002} & \msd{$41.34$}{0.8} & \msd{$-0.633$}{0.015} & \msd{$0.586$}{0.002} & \msd{$53.12$}{0.4} & \msd{$-0.237$}{0.006} & \msd{$0.583$}{0.005} & \msd{$52.87$}{1.0} & \msd{$-0.222$}{0.021} \\[0.2pt]
Claude Opus 4.8 & \msd{$0.725$}{0.002} & \msd{$40.30$}{0.8} & \msd{$-0.643$}{0.004} & \msd{$0.589$}{0.006} & \msd{$52.63$}{0.8} & \msd{$-0.254$}{0.023} & \msd{$0.605$}{0.008} & \msd{$50.94$}{1.1} & \msd{$-0.300$}{0.019} \\[0.2pt]
Claude Opus 4.6 & \msd{$0.762$}{0.002} & \msd{$35.16$}{0.5} & \msd{$-0.788$}{0.007} & \msd{$0.603$}{0.003} & \msd{$52.35$}{0.9} & \msd{$-0.282$}{0.008} & \msd{$0.598$}{0.002} & \msd{$53.32$}{0.5} & \msd{$-0.263$}{0.006} \\[0.2pt]

\rowcolor{FDLightBand}\multicolumn{10}{@{}l}{\textit{Open-weight}}\\
\rowcolor{white}\addlinespace[2pt]
GLM-5 & \msd{$0.731$}{0.006} & \msd{$38.54$}{0.6} & \msd{$-0.686$}{0.016} & \msd{$0.625$}{0.004} & \msd{$50.02$}{1.0} & \msd{$-0.372$}{0.024} & \msd{$0.632$}{0.002} & \msd{$49.96$}{0.4} & \msd{$-0.398$}{0.017} \\[0.2pt]
Qwen3.5-397B & \msd{$0.759$}{0.005} & \msd{$36.54$}{0.7} & \msd{$-0.796$}{0.021} & \msd{$0.639$}{0.004} & \msd{$48.49$}{1.2} & \msd{$-0.422$}{0.007} & \msd{$0.643$}{0.010} & \msd{$48.90$}{0.9} & \msd{$-0.441$}{0.027} \\[0.2pt]
Kimi K2.5 & \msd{$0.773$}{0.002} & \msd{$36.30$}{0.8} & \msd{$-0.837$}{0.011} & \msd{$0.637$}{0.010} & \msd{$49.37$}{2.2} & \msd{$-0.420$}{0.024} & \msd{$0.658$}{0.009} & \msd{$48.07$}{1.7} & \msd{$-0.439$}{0.005} \\[0.2pt]
MiniMax M2.5 & \msd{$0.758$}{0.005} & \msd{$35.68$}{0.9} & \msd{$-0.797$}{0.011} & \msd{$0.654$$^{\ddagger}$}{0.012} & \msd{$46.87$}{0.7} & \msd{$-0.461$}{0.031} & \msd{$0.640$}{0.007} & \msd{$47.19$}{1.6} & \msd{$-0.413$}{0.022} \\[0.2pt]
DeepSeek-V3.2 & \msd{$0.805$}{0.004} & \msd{$34.34$}{1.0} & \msd{$-0.991$}{0.017} & \msd{$0.655$}{0.012} & \msd{$47.73$}{1.9} & \msd{$-0.464$}{0.042} & \msd{$0.666$}{0.019} & \msd{$47.65$}{1.5} & \msd{$-0.508$}{0.064} \\[0.2pt]
gpt-oss-120b & \msd{$0.896$}{0.004} & \msd{$32.44$}{0.5} & \msd{$-1.378$}{0.023} & \msd{$0.699$$^{\ddagger}$}{0.010} & \msd{$41.95$}{0.6} & \msd{$-0.594$}{0.020} & \msd{$0.696$$^{\S}$}{0.012} & \msd{$43.03$}{0.7} & \msd{$-0.584$}{0.052} \\[0.2pt]
Nemotron 3 Super & \msd{$0.882$$^{\dagger}$}{0.006} & \msd{$30.84$}{0.5} & \msd{$-1.516$}{0.021} & \msd{$0.695$$^{\ddagger}$}{0.011} & \msd{$43.49$}{1.2} & \msd{$-0.743$}{0.038} & \msd{$0.688$$^{\S}$}{0.019} & \msd{$44.89$}{2.7} & \msd{$-0.720$}{0.031} \\[0.2pt]
\midrule
Uniform reference & $0.771$ & $22.9$ & \textemdash{} & $0.771$ & $22.9$ & \textemdash{} & $0.771$ & $22.9$ & \textemdash{} \\
Market reference & $0.498$ & $64.5$ & \textemdash{} & $0.498$ & $64.5$ & \textemdash{} & $0.498$ & $64.5$ & \textemdash{} \\
\bottomrule
\end{tabular*}

\endgroup
\end{table}

\subsection{Evaluation and Learning}
\label{sec:method:evaluation}

We evaluate forecasts primarily with the multiclass Brier score~\citep{brier1950verification}, and additionally report top-label accuracy, dataset-level calibration error~\citep{ece}, and market-relative measures such as Information-$\alpha$. Each episode supports trajectory-level analyses of how beliefs evolve (Appendix~\ref{sec:appendix:longitudinal:index} and \ref{sec:appendix:longitudinal:capture}). With trajectory logging, each step also retains $H_t$ and, in memory-on mode, the notebook, to which outcome-based feedback can be attached. Evaluation events use these records for benchmarking, and training events supply trajectories and feedback for learning, which we demonstrate with supervised fine-tuning in Section~\ref{sec:experiments:training}.

\section{Experiments}
\label{sec:experiments}

\subsection{Experimental Setup}
\label{sec:experiments:setup}

We evaluate 12 proprietary and open-weight models on 230 held-out events (797 event--date pairs), with four rollouts per pair. To examine the roles of research tools and cross-date memory, we compare three settings on identical scheduled forecasting tasks: \emph{no-tools}, which forecasts without external tools (Figure~\ref{fig:no_tools_prompt}); \emph{memory-free}, which provides date-restricted search, article reads, and Python tools but starts from a fresh context at each date (Figure~\ref{fig:memory_free_prompt}); and \emph{memory-on}, which uses the same tools while carrying a belief notebook across forecast dates within each rollout (Figure~\ref{fig:memory_on_prompt}). To mitigate outcome leakage, all models have a reported knowledge or training-data cutoff, or otherwise a checkpoint release date, preceding the evaluation period (Table~\ref{tab:model_knowledge_cutoff}). A complementary no-tools diagnostic finds no evidence of widespread outcome recall (Appendix~\ref{sec:appendix:knowledge_cutoffs}). Full configurations and tool budgets are provided in Appendix~\ref{sec:appendix:configurations}.

\begin{figure}[!htbp]
    \centering
    \includegraphics[width=\linewidth]{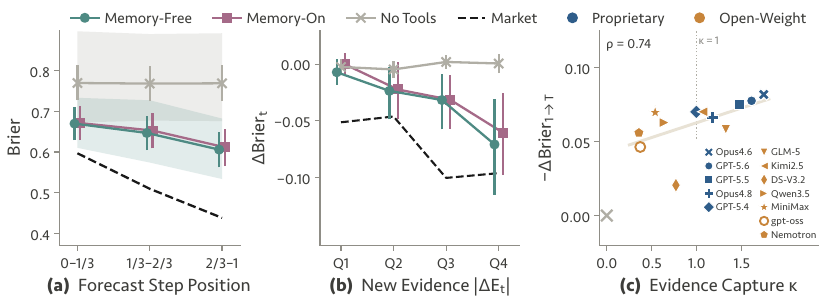}
    \caption{\textbf{Forecast improvements track newly recorded evidence.}
    (a) Brier across early, middle, and late thirds of each event trajectory, with equal event weighting; shading shows the range of model means. (b) Consecutive-step Brier change by new-evidence quartile, using leave-one-model-out evidence indices. (c) Evidence capture $\kappa$ versus first-to-last Brier improvement with memory; correlation is across 12 models, with the no-tools setting shown as a reference. Scores use the same pairing and uniform fallback as Table~\ref{tab:dojo_main_results}; (a,b) show 95\% event-bootstrap CIs. See Appendix~\ref{sec:appendix:longitudinal:capture} for definitions and sensitivity analyses.
}
    \label{fig:longitudinal_analysis}
\end{figure}

We evaluate forecasting performance using multiclass Brier score, top-label accuracy, and Information-$\alpha$, which measures the agent's log-score advantage over the market on the realized outcome (positive values favor the agent). Uniform forecasts and historical market probabilities serve as contextual references, although the market may incorporate information beyond our news archive and its probabilities are never exposed to agents. Unusable forecasts are replaced with uniform distributions over the $K$ possible outcomes, yielding a Brier score of $1-1/K$ and fractional-tie accuracy of $1/K$. Longitudinal averages weight events equally, while all 95\% confidence intervals and paired comparisons use event-level resampling (Appendix~\ref{sec:appendix:metrics}).

\subsection{Overall Forecasting Performance}
\label{sec:experiments:performance}

\paragraph{Research tools improve forecasting across model families.}
Relative to no tools, memory-free research lowers Brier and improves accuracy for all 12 models, and every paired 95\% CI excludes zero (Tables~\ref{tab:dojo_main_results} and~\ref{tab:appendix_paired}). GPT-5.5, for example, improves from 0.698 to 0.564 in Brier and from 43.85\% to 57.69\% in accuracy, and the open-weight DeepSeek-V3.2 improves from 0.805 to 0.655 and from 34.34\% to 47.73\%.

\paragraph{Proprietary models achieve lower Brier scores with tools.}
In both tool-enabled settings, all five proprietary models have lower mean Brier than every open-weight model. The proprietary group is 0.078 lower on average without memory (95\% CI $[0.056,0.103]$), although the closest pair, Claude Opus 4.6 and GLM-5, is not statistically distinguishable in a paired comparison. GPT-5.6 Sol achieves the lowest Brier without and with memory (0.554 and 0.546), and its lead over GPT-5.5 is significant only with memory (Appendix~\ref{sec:appendix:uncertainty}). GLM-5 is the strongest open-weight model (0.625 and 0.632).

\paragraph{Accuracy and Brier can disagree through calibration.}
GPT-5.5 has the highest memory-free accuracy (57.69\%) but is
underconfident. Its top-ranked outcome receives 0.50 probability on
average while being correct on 58\% of usable forecasts, the largest
top-label calibration error among the 12 models
(Appendix~\ref{sec:appendix:calibration}). GPT-5.6 Sol is slightly less
accurate (56.36\%; the paired accuracy-difference CI includes zero), partly because 4.49\% of its
memory-free runs stop before a final answer, yet it has lower Brier
(0.554 versus 0.564) and higher Information-$\alpha$
($-0.138$ versus $-0.186$).
With uniform fallback, all 12 models beat the uniform reference in
Brier under both tool-enabled settings. This convention does not imply
successful output, since 19.45\% of gpt-oss-120b's memory-free reports
are unusable (Appendix~\ref{sec:appendix:failure}).

\paragraph{Memory yields model-dependent changes in forecast quality.}
Adding a belief notebook lowers mean Brier for six models and raises
it for the other six.
None of the six improvements has a 95\% CI that excludes zero, whereas memory significantly raises Brier for Claude Opus 4.8 ($+0.016$, $[0.006, 0.025]$) and Kimi K2.5 ($+0.021$, $[0.002, 0.041]$) on all recorded pairs (Table~\ref{tab:appendix_paired}). Much of this split comes from output reliability.
For GPT-5.6 Sol, the share of forecasts without a usable answer falls from 4.49\% to 0.53\% with memory,
yet on forecasts usable in both settings the paired Brier difference is
only $+0.0010$. We therefore separate memory's effects on forecast quality and output
reliability from its effect on research cost
(Section~\ref{sec:experiments:cost_analysis}).

\paragraph{The market reference remains ahead across evaluation metrics.}
The market achieves Brier 0.498 and accuracy 64.55\%, against the best model scores of 0.546 and 57.69\%. Paired on the same event--dates, the Brier gap of the best configuration is 0.050 (95\% CI $[0.019,0.082]$), and every configuration has a strictly positive paired Brier-gap CI. Information-$\alpha$ is also negative for every model and setting, with all 95\% CIs below zero (Appendix~\ref{sec:appendix:uncertainty}). Because the market may incorporate information outside the agents' archive, it serves as an external reference rather than an information-matched baseline.

\subsection{Forecasting over Time}
\label{sec:experiments:stages}

\paragraph{Forecasts improve over time with research tools.}
Figure~\ref{fig:longitudinal_analysis}(a) compares forecasts at the early, middle, and late stages of each event.
Mean Brier decreases from 0.670 to 0.606 with memory-free research and from 0.671 to 0.612 with memory-on research.
In contrast, the no-tools baseline remains nearly unchanged near 0.769. All 12 models achieve lower mean Brier in the late than
in the early stage with memory-free research, each with
a 95\% CI below zero (Appendix~\ref{sec:appendix:longitudinal_robustness}).
The market improves more sharply, from 0.597 to 0.438. The improvement is concentrated in events on which the market also moved toward the outcome (Appendix~\ref{sec:appendix:selection_sensitivity}).
These results show that agents benefit from access to dated information as events unfold, even without retaining their previous research.

\begin{figure}[t]
    \centering
    \includegraphics[width=\linewidth]{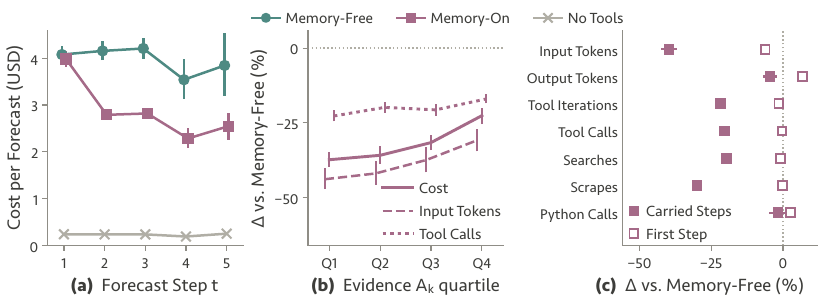}
\caption{\textbf{Memory lowers research cost, mostly after the first forecast.}
(a) Mean API cost per forecast step across five proprietary models; later steps cover fewer events (67/230 at step 4; 25/230 at step 5). (b) Consecutive-step Brier changes across quartiles of the
leave-one-model-out new-evidence index $A_k(e,t)$. (c) The same changes at the first versus later steps; notebooks are only available after the first step. Negative values denote reductions. Tokens and tool calls use all 12 models; intervals are 95\% event-bootstrap CIs.
}
    \label{fig:cost_analysis}
\end{figure}
\paragraph{Larger improvements coincide with more new evidence.}
We approximate newly available evidence using dated entries in belief notebooks. For each model, the leave-one-model-out evidence index $A_k(e,t)$ averages the number of newly dated entries recorded by other models between consecutive forecasts (Appendix~\ref{sec:appendix:longitudinal:index}). For memory-free agents, mean step-to-step Brier change ranges from $-0.007$ in the lowest index quartile to $-0.071$ in the highest (Figure~\ref{fig:longitudinal_analysis}(b)). Memory-on agents show a similar pattern, while no-tools forecasts change little across quartiles. Models recording more newly dated evidence also achieve larger episode-level improvements (Spearman $\rho=0.74$, Figure~\ref{fig:longitudinal_analysis}(c)), although this association is between models rather than within them (Appendix~\ref{sec:appendix:longitudinal:capture}).

\subsection{Cost Analysis}
\label{sec:experiments:cost_analysis}

\paragraph{Memory reduces the cost of repeated forecasting.}
For the five proprietary models, memory-on execution costs less per forecast than memory-free execution in every case, with reductions ranging from 8\% for Claude Opus 4.6 to 33\% for GPT-5.5 (median 24\%; all paired 95\% CIs exclude zero; Appendix~\ref{sec:appendix:resources}). GPT-5.5, for example, costs \$3.90 instead of \$5.85 per forecast, while its paired Brier difference of $+0.007$ has a CI that includes zero. Both settings cost about the same at the first forecast. Savings emerge from the second forecast onward, are largest when little new evidence has appeared, and coincide with fewer input tokens, tool iterations, and research calls (Figure~\ref{fig:cost_analysis}). This pattern is consistent with agents reusing earlier findings, and lower research cost is memory's clearest benefit in the evaluated protocols.

\subsection{Training Demonstration}
\label{sec:experiments:training}

We investigate whether \ours can also serve as a training environment by leveraging its collected agent interactions and outcome feedback. Using Qwen3-235B-A22B-Thinking-2507~\citep{yang2025qwen3}, we generate four memory-free trajectories for each of 3,965 event--date tasks drawn from 1,023 training events (August 2025 to February 2026). We select one demonstration per task using a weighted combination of Brier score, accuracy, and tool use, leveraging known outcomes to guide offline selection (Appendix~\ref{sec:appendix:sft}). We then fine-tune Qwen3-30B-A3B-Thinking-2507~\citep{yang2025qwen3} for three epochs using next-token cross-entropy over assistant tokens, including reasoning and tool calls (Appendix~\ref{sec:appendix:sft}).

On 230 temporally held-out events, the fine-tuned model improves both Brier score and accuracy over its base (Table~\ref{tab:dojo_sft_effect}). Brier decreases from 0.924 to 0.749, with gains on both binary and multi-choice forecasts, while accuracy rises from 34.8\% to 42.8\%. The fine-tuned model also makes more tool calls (3.7 vs.\ 1.4 per forecast). To examine the role of predictive confidence, we conduct an oracle uniform-shrinkage analysis. Rescaling the base model's probabilities reduces its Brier to 0.752, nearly matching the unadjusted SFT score of 0.749. However, shrinkage preserves top-1 predictions and therefore cannot account for the observed accuracy improvement (Appendix~\ref{sec:appendix:sft}). Together, these results demonstrate the potential of \ours as a training environment, where recorded agent trajectories can support improved forecasting on temporally held-out events.

\begin{table}[t]
\centering
\small
\setlength{\tabcolsep}{4pt}
\renewcommand{\arraystretch}{1.12}
\definecolor{Tint}{HTML}{EAF2FA}
\definecolor{Muted}{HTML}{777777}
\newcommand{\muted}[1]{\textcolor{Muted}{#1}}
\caption{\textbf{Supervised fine-tuning results.}
Paired SFT$-$base differences ($\Delta$) are computed before rounding; event-level bootstrap 95\% CIs exclude zero for Brier ($[-0.214,-0.139]$) and accuracy ($[+5.1,+11.2]$). Unusable forecasts (0\% base, 0.5\% SFT) receive the uniform fallback.}
\label{tab:dojo_sft_effect}
\vspace{-0.5em}
\begin{tabularx}{\linewidth}{l *{5}{>{\centering\arraybackslash}X}}
\toprule
 & \multicolumn{3}{c}{Brier$\downarrow$} & Acc.$\uparrow$ & Tool calls \\
\cmidrule(lr){2-4}
 & All & Binary & Multi-option & (\%) & / forecast \\
\midrule
\muted{Uniform}  & \muted{0.771} & \muted{0.500} & \muted{0.832} & \muted{22.9} & \muted{--} \\
Student (base)   & 0.924 & 0.704 & 0.975 & 34.8 & 1.4 \\
\rowcolor{Tint}
\textbf{+ Dojo SFT} & \textbf{0.749} & \textbf{0.539} & \textbf{0.797} & \textbf{42.8} & 3.7 \\
\midrule
Paired $\Delta$  & $-0.176$ & $-0.165$ & $-0.178$ & $+8.1$ & $+2.3$ \\
\bottomrule
\end{tabularx}
\end{table}

\section{Conclusion}
\label{sec:conclusion}

We introduce \ours, a replayable environment that transforms resolved prediction-market events into temporally grounded, multi-step forecasting episodes. By providing agents with identical forecast schedules and time-restricted information, \ours enables controlled evaluation of how forecasting performance, evidence use, and research efficiency evolve over time. Across 12 models, research tools improve both Brier score and accuracy, with larger forecast gains coinciding with more newly recorded evidence. Persistent memory reduces research costs more consistently than it improves forecast quality, revealing a distinction between efficient information reuse and better predictions. Beyond evaluation, \ours supports learning from recorded agent interactions: supervised fine-tuning improves student accuracy from 34.8\% to 42.8\% on temporally held-out events. Together, these findings establish \ours as a unified foundation for evaluating, understanding, and improving forecasting agents as new information unfolds.

\bibliography{ref}
\bibliographystyle{iclr2027_conference}

\appendix
\providecommand{\ours}{Forecast-Dojo}
\providecommand{\cicell}[3]{\begin{tabular}[t]{@{}c@{}}#1\\[-2pt]{\scriptsize\textcolor{black!55}{[#2,\,#3]}}\end{tabular}}

\appendix
\renewcommand{\thesection}{\Alph{section}}
\newpage

\begin{center}
    {\Large \textbf{Appendix for \ours}}
\end{center}

\startcontents[sections]
\printcontents[sections]{l}{1}{\setcounter{tocdepth}{2}}

\section{Dataset and Benchmark Details}
\label{sec:appendix:data}

We describe in detail how we construct our information corpus (\textsection\ref{sec:appendix:data:corpus}), collect and filter forecasting events (\textsection\ref{sec:appendix:data:event}), and select forecast dates for each task (\textsection\ref{sec:appendix:data:step}).

\subsection{Information Corpus}
\label{sec:appendix:data:corpus}

\paragraph{Corpus source.}
We build the retrieval corpus from the monthly CC-News releases of Common Crawl~\citep{nagel2016ccnews} from March 2025 through May 2026. For each news page, we extract the title, summary, body text, URL, crawl time, and structured publication metadata.

\paragraph{Text and quality filtering.}
Our filtering pipeline is summarized in
Table~\ref{tab:corpus_filter_config}. The first two stages normalize
the extracted text and retain English-language documents using
fastText~\citep{joulin2017bag,joulin2016fasttext}, matching the
language of the benchmark questions. The remaining stages adapt quality
filters from Gopher~\citep{gopher}, C4~\citep{c4}, and
FineWeb~\citep{fineweb}.

\begin{table}[t]
\centering
\small
\begin{tabular}{p{0.23\linewidth}p{0.7\linewidth}}
\toprule
\textbf{Stage} & \textbf{Deployed criterion} \\
\midrule

1. Cleaning &
Normalize whitespace and inline URLs; drop documents with an empty cleaned title or body. \\
2. Language &
Keep documents that fastText identifies as English with confidence $\ge 0.5$. \\
3. Gopher repetition &
Apply duplicate-line, duplicate-paragraph, and repeated $n$-gram checks. \\
4. Gopher quality &
Require 50--100,000 non-symbol tokens, mean word length in $[3,10]$, bounded symbol, bullet, and ellipsis ratios, alphabetic-word ratio $\ge 0.75$, and at least two stop-word occurrences. \\
5. C4 quality &
Apply C4 line and document cleanup and require at least five sentences. \\
6. FineWeb quality &
Require at least $12\%$ of lines to end with terminal punctuation and at most $80\%$ short lines. \\
7a. URL deduplication &
Remove documents whose canonicalized URL has already been seen.\\
7b. Title deduplication &
Suppress identical normalized titles appearing within a seven-day window, retaining the earliest-published document. \\
7c. MinHash deduplication &
Apply MinHash-LSH over word 5-grams and retain the earliest-published representative of each near-duplicate component. \\
8. Serving-date window &
Retain documents whose resolved publication date falls within March 1, 2025 through May 31, 2026. \\
\bottomrule
\end{tabular}

\caption{
Corpus filtering pipeline. The aggregate post-quality
count is reported in Table~\ref{tab:filter_summary:corpus}.
}
\label{tab:corpus_filter_config}
\end{table}

\paragraph{Publication dates and temporal integrity.}
We read publication and modification timestamps from structured page metadata~\citep{guha2016schema,schemaDatePublished,googleArticleStructuredData}, such as schema.org and OpenGraph fields, and discard values outside 2010--2030. Each article receives the UTC day of the structured timestamp closest to its crawl time, preferring the later one in case of a tie and never exceeding the crawl time. Articles without a valid structured timestamp receive their crawl day. Across the corpus, the assigned date comes from publication metadata for $58.6\%$ of articles, from modification metadata for $28.3\%$, and from the crawl time for $13.1\%$.

Retrieval at forecast date $\tau_t$ includes articles assigned to $\tau_t$ itself. Because dates are assigned at day precision, they do not show that a page existed at a particular time of day or that an edited page already had its captured content on the assigned date. The leakage audit in Appendix~\ref{sec:appendix:data:event} addresses the violations we observed.

\paragraph{Duplicate suppression.}
We remove repeated coverage by URL matching, suppression of identical titles within a seven-day window, and MinHash near-duplicate detection over word 5-grams, keeping the earliest-published copy in each case. When a URL was captured more than once, we keep a single capture, so a retained page is not necessarily its earliest version.

\paragraph{Indexing and temporal restriction.}
Each retained article is embedded with Qwen3-Embedding-8B~\citep{qwen3embedding} using its title, summary, and body. The normalized embeddings are stored in exact inner-product indices, one per month. At forecast date $\tau_t$, search is restricted to
\[
\mathcal I_{\leq\tau_t}
=
\{d\in\mathcal I:
2025\text{-}03\text{-}01
\leq t(d)
\leq \tau_t\}.
\]
The date restriction is applied before similarity ranking, and full-article access independently rechecks the same temporal constraint.

\subsection{Event Collection and Filtering}
\label{sec:appendix:data:event}

\paragraph{Event collection and market beliefs.}
We construct forecasting questions from resolved Polymarket events.
A binary event is a single yes/no market, and a multi-option event is a group of at least two mutually exclusive markets. Other market layouts are excluded. For market
leg $k$, let $t_k^{\mathrm{start}}$ and $t_k^{\mathrm{close}}$ denote
its recorded start and close times. We define the complete tradeable
span as
\[
[t_{\mathrm{start}},t_{\mathrm{close}}]
=
\left[
\min_k t_k^{\mathrm{start}},
\max_k t_k^{\mathrm{close}}
\right].
\]
Ground-truth labels are derived from terminal YES prices: binary markets
resolve to YES at $\ge 0.99$ and to NO at $\le 0.01$, while multi-option
events require exactly one YES-resolved option. We reconstruct the daily
market belief $m_{Q,t}\in\Delta^{K-1}$ from the UTC-day mean of YES trade prices. For a multi-option event with priced options
$A_t$,
\[
m_{Q,t}(k)
=
\frac{x_{t,k}}
{\sum_{j\in A_t} x_{t,j}},
\qquad k\in A_t,
\]
with missing options left undefined rather than imputed.

\paragraph{Market-side filtering.}
We first filter events for a meaningful forecasting horizon and a reliable, nontrivial market-belief trajectory. Events must span at least eight UTC calendar days. We then define a day as \emph{active} when the trade count and share volume on the market leg corresponding to the realized outcome exceed the type-specific thresholds in Table~\ref{tab:event_filter_criteria}. Let $\rho_Q$ denote the fraction of active days and $S_Q$ the longest consecutive run of inactive days.

Among events passing this activity filter, we measure task nontriviality
from the market probability assigned to the realized outcome, $m_{Q,t}(Y)$. Let $\mathcal F_Q$ denote active days with an observed belief, and let $\Delta_t=m_{Q,t}(Y)-m_{Q,t-1}(Y)$ for adjacent calendar days with valid active-day observations. We define
\[
a_Q
=
\frac{1}{|\mathcal F_Q|}
\sum_{t\in\mathcal F_Q}
-\log
\operatorname{clip}
\left(m_{Q,t}(Y),10^{-6},1-10^{-6}\right),
\quad
V_Q=\sum_t|\Delta_t|,
\quad
C_Q=\frac{\sum_t\Delta_t}{\sum_t|\Delta_t|}.
\label{eq:event_filter_signal}
\]
These remove events that are already nearly certain or have
little meaningful temporal variation. 

\begin{table}[t]
\centering
\small
\begin{tabular}{p{0.30\linewidth}p{0.28\linewidth}p{0.28\linewidth}}
\toprule
\textbf{Criterion} &
\textbf{Binary} &
\textbf{Multi-option} \\
\midrule

Forecasting horizon &
$\ge 8$ days &
$\ge 8$ days \\

Active day &
$\ge 6$ trades,
$\ge 200$ shares &
$\ge 5$ trades,
$\ge 50$ shares \\

Active-day coverage &
$\rho_Q \ge 0.80$ &
$\rho_Q \ge 0.70$ \\

Max.\ inactive run &
$S_Q \le 3$ &
$S_Q \le 5$ \\

Residual uncertainty &
$a_Q \ge 0.05$ &
$a_Q \ge 0.05$ \\

Temporal signal &
\multicolumn{2}{c}{
$V_Q\ge10^{-6}$ and
$(C_Q\ge0.20 \lor V_Q\ge0.50)$
} \\

\bottomrule
\end{tabular}
\caption{Market-side event-selection criteria.}
\label{tab:event_filter_criteria}
\end{table}

\paragraph{Corpus-side filtering.}
Market-side filtering identifies events with usable forecasting trajectories, but does not establish whether the frozen information environment is suitable for forecasting. We therefore apply a complementary corpus-side screen that evaluates both the sufficiency of pre-resolution evidence and potential information leakage.

\noindent\textit{Evidence sufficiency.}
Following the sufficient-context framework of \citet{joren2024sufficient}, we assess whether the retrieved information contains enough evidence to support an informed forecast. We adapt this protocol to historical forecasting by first decomposing each event into targeted information needs. For each event surviving the market-side filters, we use Claude Opus 4.8 with maximum reasoning effort~\citep{anthropic2026opus48} to generate 4--7 predictive-evidence queries and 4--7 background/reference-class queries. Query generation explicitly targets information available before resolution and forbids searches for the realized outcome or post-resolution reports. See the full decomposition prompt in Figure~\ref{fig:query_decomposition_prompt}.

Sufficiency is evaluated once at the latest eligible forecasting state,
\[
\tau^\star=t_{\mathrm{close}}-2\ \text{days}.
\]
The buffer is relative to the recorded market close. It reduces exposure to close-day reporting and adjacent-day timestamp ambiguity, but does not ensure that the deciding event was still unresolved. For each query, we retrieve dense (Qwen3-Embedding-8B) and BM25~\citep{bm25} top-10 results restricted to $[2025\text{-}03\text{-}01,\tau^\star]$ and combine them using reciprocal-rank fusion,
\[
\operatorname{RRF}(d)
=
\sum_q
\sum_{r\in\{\mathrm{dense},\mathrm{BM25}\}}
\frac{\mathbf{1}[d\in L_{q,r}]}
{60+\operatorname{rank}_{q,r}(d)}.
\]
The 15 highest-ranked unique documents form the evidence set $E_Q$.
We then use Claude Opus 4.6~\citep{anthropic2026claudeopus46} to classify the available evidence as \{\textsc{Sufficient}, \textsc{Partial}, \textsc{Insufficient}\}. Only events receiving a \textsc{Sufficient} verdict are eligible for the final benchmark.

\noindent\textit{Leakage control.}
The same judge additionally screens the retrieved evidence for
potential outcome leakage. This content-level check complements the
metadata-level temporal restriction on retrieval and provides an
additional safeguard when the visible content of a page may not be
fully reflected by its assigned publication timestamp. The judge does
not receive the structured realized outcome, final market prices, or
crowd trajectory.

Let $v_Q$ denote the evidence-sufficiency verdict and $\ell_Q$ the
leakage flag. At this screening stage, an event is eligible iff
\[
v_Q=\textsc{Sufficient}
\qquad\text{and}\qquad
\ell_Q=\mathrm{false}.
\]
See Figure~\ref{fig:evidence_judge_prompt} for the judge prompt and Table~\ref{tab:filter_summary} for event filtering funnels.

\begin{table*}[t]
\centering
\caption{
Construction waterfalls for the deployed news corpus and forecasting
events. Event counts use the final June--March training and March--June evaluation splits.
}
\begin{subtable}[t]{0.48\textwidth}
\centering
\caption{Corpus construction}
\label{tab:filter_summary:corpus}
\scriptsize
\begin{tabular}{@{}p{0.32\linewidth}p{0.24\linewidth}p{0.26\linewidth}@{}}
\toprule
\textbf{Stage} & \textbf{Removed} & \textbf{Remaining} \\
\midrule
Stages 1--6 & -- & 26,533,747 \\
Stage 7a & 194,940 & 26,338,807 \\
Stage 7b & 5,563,879 & 20,774,928 \\
Stage 7c & 1,124,734 & 19,650,194 \\
Stage 8 & 823,254 & \textbf{18,826,940} \\
\bottomrule
\end{tabular}
\end{subtable}
\hfill
\begin{subtable}[t]{0.48\textwidth}
\centering
\caption{Event selection}
\label{tab:filter_summary:event}
\scriptsize
\begin{tabular}{@{}p{0.48\linewidth}r r@{}}
\toprule
\textbf{Stage} & \textbf{Train} & \textbf{Eval} \\
\midrule
Temporal split pool & 13,671 & 13,920 \\
  Forecasting horizon & 9,051 & 7,590 \\
  Market activity & 1,965 & 423 \\
  Task nontriviality & 1,777 & 378 \\
  Evidence sufficiency \& judge & \textbf{1,338} & \textbf{230} \\
\bottomrule
\end{tabular}
\end{subtable}

\label{tab:filter_summary}
\end{table*}

\begin{table}[t]
\centering
\small
\caption{Tools details.}
\label{tab:agent_tools}
\begin{tabularx}{\linewidth}{@{}lX@{}}
\toprule
\textbf{Tool} & \textbf{Semantics and permissions} \\
\midrule
\texttt{search} & Dense retrieval over the assigned-date window, with default $k=5$. Returns article IDs, titles, URLs, assigned dates, similarity scores, and short snippets. Candidate dates are restricted before ranking. \\
\texttt{scrape} & Returns the article body for an ID returned by search in the same forecast step, after rechecking the date boundary. Bodies are truncated to 8,000 tokens. \\
\texttt{python} & Executes code for numerical analysis in an isolated runtime with networking disabled and no access to benchmark data files, including the task files, news corpus, and retrieval index. \\
\bottomrule
\end{tabularx}
\label{tab:tools}
\end{table}

\subsection{Forecast-Step Construction}
\label{sec:appendix:data:step}

\paragraph{Temporal split.}
We split at the event level using the complete tradeable lifetime. An event enters training iff
\(
t_{\mathrm{start}}\ge 2025\text{-}06\text{-}01
\quad\text{and}\quad
t_{\mathrm{close}}<2026\text{-}03\text{-}01,
\)
and enters evaluation iff
\(
t_{\mathrm{start}}\ge 2026\text{-}03\text{-}01
\quad\text{and}\quad
t_{\mathrm{close}}<2026\text{-}06\text{-}01.
\)
Events that cross the March 1 boundary are excluded from both splits.

\paragraph{Number of forecast steps.}
For event $Q$, let $D_Q$ be its contiguous UTC daily grid. We reserve a two-day buffer before the recorded close date and define
\[
E_Q
=
\{d\in D_Q:\operatorname{date}(t_{\mathrm{close}})-d\ge2\},
\qquad n_Q=|E_Q|.
\]
The number of forecast steps is
\[
T_Q
=
\operatorname{clamp}
\left(
\operatorname{round}\sqrt{n_Q},
3,
10
\right).
\label{eq:appendix_num_steps}
\]

\paragraph{Date selection.}
Within the eligible grid, market movement is
\[
b_t=
\begin{cases}
|m_{Q,t}(\mathrm{YES})-m_{Q,t-1}(\mathrm{YES})|, & \text{binary},\\[2mm]
\frac12\sum_k|m_{Q,t}(k)-m_{Q,t-1}(k)|, & \text{multi-option},
\end{cases}
\]
with zero assigned when an adjacent belief is unavailable. Corpus activity $c_t$ is the total number of indexed news articles published on calendar day $t$. It is a global news-volume signal rather than an event-specific relevance score. After independently max-normalizing both signals, eligible date $e_j$ receives
\[
s_j
=
0.7\,\tilde b_{e_j}
+
0.3\,\tilde c_{e_j}
+
10^{-3}\frac{j}{n_Q-1}.
\]
We partition the eligible sequence into $T_Q$ contiguous equal-count bins and choose the highest-scoring date from each bin. This preserves temporal coverage while favoring dates with larger belief changes or greater overall news activity.

\section{Experiment Details}
\label{sec:appendix_exp}

\subsection{Knowledge Cutoffs and Checkpoint Dates}
\label{sec:appendix:knowledge_cutoffs}
Table~\ref{tab:model_knowledge_cutoff} lists each model's reported knowledge cutoff or checkpoint date. All dates precede the evaluation period, which begins on March 1, 2026.

\paragraph{Diagnostic for outcome recall.}
To further assess potential outcome recall beyond the knowledge-cutoff checks, we examine models' no-tools forecasting performance directly. If models frequently recalled resolved outcomes, they would be expected to perform well even when the contemporaneous market favored an incorrect outcome. The market favorite was wrong on 35.4\% of evaluation dates. On these dates, no-tools accuracy ranged from 11.4\% to 16.9\% across the 12 models, below the 19.1\% accuracy of a uniform forecast, compared with 41.5\% to 63.3\% when the market favorite was correct. The same pattern holds for GPT-5.6 Sol, whose reported cutoff is closest to the evaluation period (16.8\% versus 63.3\%). Moreover, no-tools Information-$\alpha$ at the first forecast date is negative for every model. Together, these findings provide no evidence of widespread usable outcome recall, although they cannot rule out partial memorization or other post-cutoff knowledge.

\begin{table}[t]
    \centering
    \caption{
        Knowledge and release dates of evaluated models.
        We report an official knowledge or training-data cutoff when available;
        otherwise, we use the public checkpoint release date.
    }
    \label{tab:model_knowledge_cutoff}
    \small
    \setlength{\tabcolsep}{4.5pt}
    \renewcommand{\arraystretch}{1.08}

    \begin{tabularx}{\columnwidth}{
        @{}
        >{\raggedright\arraybackslash}X
        >{\centering\arraybackslash}p{1.75cm}
        >{\raggedleft\arraybackslash}p{1.65cm}
        @{}
    }
        \toprule
        \textbf{Model}
        & \textbf{Date}
        & \textbf{Basis} \\
        \midrule

        \rowcolor{black!6}
        \multicolumn{3}{@{}l}{\textit{Proprietary}} \\

        GPT-5.6 Sol       & 2026-02-16 & Cutoff \\
        GPT-5.5           & 2025-12-01 & Cutoff \\
        GPT-5.4           & 2025-08-31 & Cutoff \\
        Claude Opus 4.8   & 2026-01    & Cutoff \\
        Claude Opus 4.6   & 2025-05    & Cutoff \\

        \addlinespace[2pt]
        \rowcolor{black!6}
        \multicolumn{3}{@{}l}{\textit{Open-weight}} \\

        GLM-5              & 2026-02-11 & Release \\
        Qwen3.5-397B       & 2026-02-16 & Release \\
        Kimi K2.5          & 2026-01-27 & Release \\
        MiniMax M2.5       & 2026-02-12 & Release \\
        DeepSeek-V3.2      & 2025-12-01 & Release \\
        gpt-oss-120b       & 2024-06-01 & Cutoff \\
        Nemotron 3 Super   & 2026-02    & Post-train \\

        \bottomrule
    \end{tabularx}

    \vspace{2pt}
    \begin{minipage}{\columnwidth}
        \footnotesize
        \textit{Note.}
        ``Release'' denotes the public checkpoint release date when no
        official knowledge cutoff is reported.
    \end{minipage}
\end{table}

\subsection{Models and Execution Protocols}
\label{sec:appendix:configurations}
Table~\ref{tab:model_configs} lists the decoding settings and tool budgets of the 12 model configurations. All runs use the same 230 held-out events, four rollouts per date. Tool runs share a budget of 120 tool iterations and 400 calls per forecast step. No-tool runs remove the tools and the corpus and raise the output cap to 65{,}536 tokens, since the whole forecast is then produced in a single call. Settings for the training use case are given in Appendix~\ref{sec:appendix:sft}.
\providecommand{\cfgdefault}{\textcolor{black!45}{default}}
\begin{table}[t]
    \centering
    \footnotesize
    \setlength{\tabcolsep}{5pt}
    \renewcommand{\arraystretch}{1.08}
    \caption{\textbf{Model Evaluation Configurations.} Max output tokens is the cap per model call and includes reasoning tokens. \cfgdefault{} denotes the provider's default value. Tool budgets are per forecast step and identical in memory-free and memory-on mode; no-tool runs disable all tools.}
    \label{tab:model_configs}
    \begin{threeparttable}
    \begin{tabular}{@{}llrrccc@{}}
        \toprule
        & & \multicolumn{2}{c}{Max output tokens} & \multicolumn{2}{c}{Temperature / top-$p$ / top-$k$} & Tool budget \\
        \cmidrule(lr){3-4}\cmidrule(lr){5-6}
        Model & Reasoning & Tools & No tools & Tools & No tools & iters / calls \\
        \midrule
        GPT-5.6 Sol    & max    & 32{,}768 & 65{,}536 & \cfgdefault & \cfgdefault & 120 / 400 \\
        GPT-5.5          & xhigh  & 32{,}768 & 65{,}536 & \cfgdefault & \cfgdefault & 120 / 400 \\
        GPT-5.4          & xhigh  & 32{,}768 & 65{,}536 & \cfgdefault & \cfgdefault & 120 / 400 \\
        Opus 4.8         & max    & 32{,}768 & 65{,}536 & \cfgdefault & \cfgdefault & 120 / 400 \\
        Opus 4.6         & max    & 32{,}768 & 65{,}536 & \cfgdefault & \cfgdefault & 120 / 400 \\
        GLM-5            & high   & 32{,}768 & 65{,}536 & 1.0 / 1.0 / \cfgdefault & 1.0 / 1.0 / \cfgdefault & 120 / 400 \\
        Kimi K2.5        & high   & 32{,}768 & 65{,}536 & 1.0 / 1.0 / \cfgdefault & 1.0 / 1.0 / \cfgdefault & 120 / 400 \\
        DeepSeek V3.2    & high   & 32{,}768 & 65{,}536 & 1.0 / 1.0 / \cfgdefault & 1.0 / 1.0 / \cfgdefault & 120 / 400 \\
        MiniMax M2.5     & native\tnote{a} & 32{,}768 & 65{,}536 & 1.0 / 1.0 / \cfgdefault & 1.0 / 1.0 / \cfgdefault & 120 / 400 \\
        Nemotron 3 Super & high   & 32{,}768 & 65{,}536 & 1.0 / 1.0 / \cfgdefault & 1.0 / 1.0 / \cfgdefault & 120 / 400 \\
        gpt-oss-120b     & high   & 32{,}768 & 65{,}536 & 1.0 / 1.0 / \cfgdefault & 1.0 / 1.0 / \cfgdefault & 120 / 400 \\
        Qwen3.5-397B     & native\tnote{a} & 32{,}768 & 65{,}536 & 1.0 / 1.0 / \cfgdefault & 1.0 / 1.0 / \cfgdefault & 120 / 400 \\
        \bottomrule
    \end{tabular}
    \begin{tablenotes}[flushleft]
        \footnotesize
        \item[a] The model reasons by default and exposes no effort setting.
    \end{tablenotes}
    \end{threeparttable}
\end{table}

\subsection{Evaluation Metrics}
\label{sec:appendix:metrics}

\paragraph{Brier score.}
Let $i=(e,t,r)$ index an event, forecast date, and rollout, and let $Y_i$ denote the realized outcome. For a usable probability report, negative entries are clipped to zero and the remaining positive mass is normalized. Reported labels must match the offered outcome labels exactly. Omitted outcomes receive zero probability, and probability placed on a label that matches no offered outcome counts as error. Let $\mathcal U_i$ be the union of the reported labels and the truth label. The multiclass Brier score is
\[
 B_i=\sum_{c\in\mathcal U_i}
 \left(p_{ic}-\mathbf 1\{c=Y_i\}\right)^2.
 \label{eq:eval:brier}
\]
We use the standard $[0,2]$ scale, without normalization by the number of outcomes or an additional binary-event factor. A uniform forecast over $K_i$ offered outcomes has Brier score $1-1/K_i$.

\paragraph{Accuracy.}
Let $\mathcal T_i=\arg\max_c p_{ic}$ denote the set of outcomes assigned maximal probability. We use fractional-tie accuracy,
\[
 a_i=
 \frac{\mathbf 1\{Y_i\in\mathcal T_i\}}
 {|\mathcal T_i|}.
 \label{eq:eval:accuracy}
\]
Thus, a correct unique top prediction receives accuracy $1$, while ties split credit uniformly among tied outcomes. A uniform forecast over $K_i$ outcomes therefore has accuracy $1/K_i$.

\paragraph{Information-$\alpha$.}
We measure improvement over the contemporaneous market belief using
\[
 \alpha_i=
 \log\max\{p_i(Y_i),\epsilon\}
 -
 \log\max\{p_{\mathrm{market},i}(Y_i),\epsilon\},
 \qquad \epsilon=10^{-3},
 \label{eq:eval:infoalpha}
\]
with natural logarithms; positive values favor the agent. We compute it for every recorded forecast whose market probability for the realized outcome is available, using the uniform fallback for unusable reports. Neither realized outcomes nor market probabilities are provided to the forecasting agent.

\paragraph{Log loss.}
We report $\ell_i=-\log\max\{p_i(Y_i),10^{-3}\}$ with natural logarithms. An unusable report contributes $\log K_i$.

\paragraph{Top-label calibration.}
For a usable forecast, let $c_i=\max_c p_{ic}$ be its top-ranked probability. We group forecasts into 15 equal-width bins of $c_i$ and report $\mathrm{ECE}=\sum_b \frac{n_b}{n}\,\lvert\bar c_b-\bar a_b\rvert$, where $\bar c_b$ and $\bar a_b$ are the mean top-ranked probability and the mean fractional-tie accuracy in bin $b$.

\paragraph{Statistical uncertainty.}
We account for dependencies among repeated forecasts of the same event by computing 95\% confidence intervals using percentile bootstrapping over the 230 evaluation events (4,000 resamples; seed 0). For paired comparisons, we average the four rollouts within each event--date before computing matched differences, except for memory comparisons, which preserve event--date--rollout pairing. Bootstrap resampling is performed at the event level throughout. The $\pm$ values in Table~\ref{tab:dojo_main_results} instead report the standard deviation across four rollout means, reflecting run-to-run variability rather than event-level uncertainty.

\subsection{Forecast Validity and Failure Handling}
\label{sec:appendix:failure}

To ensure consistent scoring across models and evaluation settings, we assign a uniform forecast to any recorded output that cannot be reliably scored, including incomplete reports, formatting violations (Appendix E), and predictions without positive finite probability mass. For an event with $K_i$ possible outcomes, this yields a Brier score of $1-1/K_i$, fractional-tie accuracy of $1/K_i$, and log loss of $\log K_i$. This convention applies consistently across the main evaluation, longitudinal analyses, and SFT experiments, while Information-$\alpha$ is computed only when the corresponding market probability is available.

Missing records are distinguished from unusable outputs: they are excluded from metric averages but included in the overall failure rates reported in Table~\ref{tab:failure_rates}. Of the 3,188 scheduled forecasts per configuration, only two memory-on configurations have missing records: GPT-5.6 Sol (3,184 recorded) and Nemotron 3 Super (3,170 recorded). All other configurations have complete records.

\begin{table*}[t]
\centering
\footnotesize
\setlength{\tabcolsep}{4pt}
\renewcommand{\arraystretch}{1.12}
\caption{\textbf{Forecast failure rates and causes across evaluation settings (\%).}
Early denotes termination before a final answer due to context, token, or tool limits; Format denotes an unscorable final report. Total includes missing or unusable forecasts, including invalid probability mass, over 3,188 scheduled forecasts per model and setting. Early and Format use recorded forecasts as the denominator and therefore need not sum to Total. Recorded unusable forecasts are scored as uniform, while missing records are excluded from score means.}
\label{tab:failure_rates}
\label{tab:appendix_failure_types}
\vspace{-0.5em}
\begin{tabular*}{\textwidth}{@{\extracolsep{\fill}}l*{9}{r}@{}}
\toprule
& \multicolumn{3}{c}{No tools}
& \multicolumn{3}{c}{Memory-free}
& \multicolumn{3}{c}{Memory-on} \\
\cmidrule(lr){2-4}
\cmidrule(lr){5-7}
\cmidrule(lr){8-10}
Model
& Early & Format & Total
& Early & Format & Total
& Early & Format & Total \\
\midrule
GPT-5.6 Sol
& 0.00 & 0.03 & 0.03
& 4.49 & 0.00 & 4.49
& 0.41 & 0.00 & 0.53 \\
GPT-5.5
& 0.00 & 0.06 & 0.06
& 0.16 & 0.09 & 0.25
& 0.00 & 0.03 & 0.03 \\
GPT-5.4
& 0.00 & 0.00 & 0.00
& 0.09 & 0.00 & 0.09
& 0.00 & 0.00 & 0.00 \\
Claude Opus 4.8
& 0.00 & 0.00 & 0.00
& 0.00 & 0.00 & 0.00
& 0.00 & 0.00 & 0.00 \\
Claude Opus 4.6
& 0.38 & 0.00 & 0.38
& 0.00 & 0.03 & 0.03
& 0.00 & 0.00 & 0.00 \\
GLM-5
& 0.00 & 0.28 & 0.28
& 0.00 & 1.47 & 1.47
& 0.00 & 1.13 & 1.13 \\
Kimi K2.5
& 0.00 & 0.00 & 0.00
& 0.03 & 2.13 & 2.16
& 0.00 & 2.45 & 2.45 \\
DeepSeek V3.2
& 0.00 & 0.25 & 0.25
& 0.03 & 1.73 & 1.76
& 0.00 & 1.88 & 1.88 \\
MiniMax M2.5
& 0.00 & 0.63 & 0.63
& 0.03 & 5.49 & 5.52
& 0.00 & 1.85 & 1.85 \\
Nemotron 3 Super
& 0.00 & 9.16 & 9.16
& 10.35 & 7.72 & 18.07
& 2.37 & 8.52 & 11.39 \\
gpt-oss-120b
& 0.00 & 0.00 & 0.00
& 0.00 & 19.45 & 19.45
& 0.00 & 13.64 & 13.64 \\
Qwen3.5-397B
& 0.00 & 0.03 & 0.03
& 0.00 & 1.10 & 1.10
& 0.00 & 1.16 & 1.16 \\
\bottomrule
\end{tabular*}
\end{table*}

\paragraph{Failure characteristics.}
Table~\ref{tab:failure_rates} further distinguishes premature termination from completed but unscorable reports. Early termination is uncommon overall, occurring in 0.03\% of no-tools, 1.27\% of memory-free, and approximately 0.23\% of memory-on forecasts. In the memory-free setting, early stops arise primarily from output-token limits (0.52\%), tool-iteration limits (0.41\%), and context overflow (0.31\%), with tool-call limits accounting for another 0.03\%. Failure patterns nevertheless vary substantially across models. For GPT-5.6 Sol, early termination accounts for all unusable memory-free forecasts (4.49\%), falling to 0.41\% with memory enabled. In contrast, gpt-oss-120b's failures arise entirely from completed but unscorable reports, affecting 19.45\% and 13.64\% of memory-free and memory-on forecasts, respectively.

\subsection{Inference Cost and Tool Usage}
\label{sec:appendix:resources}

\begin{figure*}[t]
    \centering
    \includegraphics[width=\textwidth]{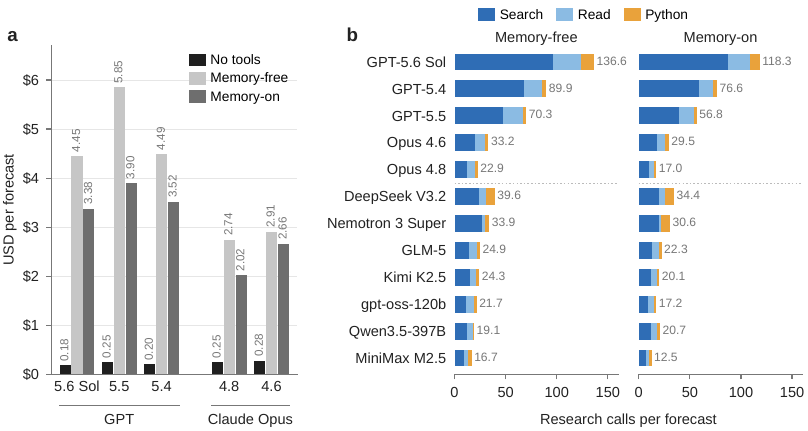}
    \vspace{-0.5em}
    \caption{\textbf{Inference cost and research tool usage across models.}
    (a) Average provider-estimated inference cost per recorded forecast under no-tools, memory-free, and memory-on settings for five proprietary models with available pricing.
    (b) Average research calls per recorded forecast, decomposed into search, full-article reads, and Python calls, including unsuccessful calls. Open-weight inference costs are omitted because comparable pricing is unavailable.}
    \label{fig:cost_details}
\end{figure*}

Figure~\ref{fig:cost_details} characterizes the inference cost and research activity of forecasting agents across evaluation settings. Panel (a) reports average provider-estimated costs for the five proprietary models with available pricing, while panel (b) breaks down research calls into search, full-article reads, and Python across all 12 models. Tool-enabled forecasting incurs substantially higher inference costs than no-tools forecasting, whereas carrying a belief notebook reduces costs for all five priced models and research calls for 11 of the 12 models.

To quantify these savings, we compute resource reductions over matched model--event--date--rollout records using ratios of summed resources, with 95\% confidence intervals from 4,000 event-level bootstrap resamples. Memory-on reduces inference cost by 23.9\% $[20.6, 27.3]$ for GPT-5.6 Sol, 33.3\% $[31.4, 35.1]$ for GPT-5.5, and 21.5\% $[19.2, 23.8]$ for GPT-5.4. Corresponding reductions are 26.2\% $[23.8, 28.6]$ for Claude Opus 4.8 and 8.3\% $[6.2, 10.4]$ for Claude Opus 4.6. These estimates characterize provider-reported costs and observed tool usage rather than hardware-normalized efficiency; comparable inference costs are unavailable for the open-weight models.

\subsection{Measuring Newly Recorded Evidence}
\label{sec:appendix:longitudinal:index}

\paragraph{Longitudinal evaluation.}
To characterize forecasting performance over time, we assign
each forecast step $t$ of event $e$ a normalized position
$(t-1)/(T_e-1)$. We divide these positions into early,
middle, and late stages using the right-closed intervals
$[0,1/3]$, $(1/3,2/3]$, and $(2/3,1]$, respectively.
Brier scores are averaged within each event and stage before
aggregation with equal event weights. Figure~\ref{fig:longitudinal_analysis}(a)
uses 223 events with agent observations across all three stages
(224 for the market). We compute 95\% confidence intervals
using 4,000 event-level bootstrap resamples.

\paragraph{Newly dated evidence.}
To examine how forecast improvements relate to newly available information, we measure evidence recorded in memory-on belief notebooks. For model $k$, rollout $r$, event $e$, and forecast step $t\geq2$, we define
\[
n_{k,r}(e,t)
=
\sum_{a\in\operatorname{ledger}(M_{k,r}(e,t))}
\mathbf 1\!\left\{
\tau_{e,t-1}<d(a)\leq\tau_{e,t}
\right\},
\]
where $d(a)$ denotes the evidence date recorded in entry $a$. This counts notebook entries dated between consecutive forecasts, including active and superseded entries, while excluding invalid or future-dated evidence. Earlier evidence discovered later and entries carried over from previous steps do not contribute.

\paragraph{Leave-one-model-out evidence index.}
To avoid directly relating a model's forecast changes to its own evidence-recording behavior, we construct a leave-one-model-out index. We first average evidence counts over available rollouts of each model:
\[
\operatorname{own}_k(e,t)
=
\frac{1}{|\mathcal R_k(e,t)|}
\sum_{r\in\mathcal R_k(e,t)}
n_{k,r}(e,t),
\]
where $\mathcal R_k(e,t)$ denotes the available rollouts. We then average across the other models with available counts:
\[
A_k(e,t)
=
\frac{1}{|\mathcal P_k(e,t)|}
\sum_{j\in\mathcal P_k(e,t)}
\operatorname{own}_j(e,t),
\]
where $\mathcal P_k(e,t)$ excludes model $k$.
We refer to $A_k(e,t)$ as the \emph{new-evidence index} and use it as a proxy for newly available evidence when analyzing model $k$, including its memory-free forecasts.

For evaluation setting $s\in\{\mathrm{NT},\mathrm{MF},\mathrm{MO}\}$, denoting no tools, memory-free, and memory-on, respectively, let $B^{(s)}_{k,r}(e,t)$ be the Brier score of model $k$ at event-step $(e,t)$ in rollout $r$. We measure consecutive-step changes as
\[
\Delta B^{(s)}_{k,r}(e,t)
=
B^{(s)}_{k,r}(e,t)
-
B^{(s)}_{k,r}(e,t-1),
\]
where negative values indicate improvement. Figure~\ref{fig:longitudinal_analysis}(b) groups transitions by quartiles of $A_k(e,t)$, averaging first within events and then across events, with event-level bootstrap confidence intervals.

The index reflects newly dated evidence recorded by other agents rather than all available information, potentially missing relevant evidence that none retrieves. Table~\ref{tab:appendix_quartiles} provides the quartile-level results underlying Figure~\ref{fig:longitudinal_analysis}(b). Tool-enabled forecasts show progressively larger Brier reductions in higher-evidence quartiles, while no-tools forecasts exhibit little change. Market forecasts also improve across all quartiles, with larger gains in the upper two. These patterns suggest that the index captures periods of greater informational activity, although they do not establish a causal relationship between new evidence and forecasting improvement.

\begin{table*}[t]
\centering
\small
\setlength{\tabcolsep}{5pt}
\renewcommand{\arraystretch}{1.15}
\caption{\textbf{Consecutive-step Brier change by new-evidence quartile.}
Values correspond to Figure~\ref{fig:longitudinal_analysis}(b).
Quartile thresholds are 1.49, 2.45, and 3.98 newly dated entries.
Brackets indicate 95\% event-level bootstrap confidence intervals;
negative values indicate improvement.
The final column reports the number of events contributing
memory-free transitions.}
\label{tab:appendix_quartiles}
\vspace{-0.5em}
\begin{tabular*}{\textwidth}{@{\extracolsep{\fill}}lccccc@{}}
\toprule
Quartile
& No tools
& Memory-free
& Memory-on
& Market
& Events \\
\midrule
Q1
& \cicell{$-0.002$}{$-0.010$}{$+0.004$}
& \cicell{$-0.007$}{$-0.018$}{$+0.005$}
& \cicell{$-0.000$}{$-0.010$}{$+0.010$}
& \cicell{$-0.051$}{$-0.095$}{$-0.013$}
& 124 \\
Q2
& \cicell{$-0.005$}{$-0.012$}{$+0.003$}
& \cicell{$-0.024$}{$-0.049$}{$-0.002$}
& \cicell{$-0.022$}{$-0.048$}{$+0.002$}
& \cicell{$-0.046$}{$-0.086$}{$-0.006$}
& 146 \\
Q3
& \cicell{$+0.002$}{$-0.004$}{$+0.008$}
& \cicell{$-0.032$}{$-0.057$}{$-0.009$}
& \cicell{$-0.031$}{$-0.057$}{$-0.009$}
& \cicell{$-0.100$}{$-0.145$}{$-0.056$}
& 135 \\
Q4
& \cicell{$+0.001$}{$-0.008$}{$+0.009$}
& \cicell{$-0.071$}{$-0.115$}{$-0.031$}
& \cicell{$-0.060$}{$-0.098$}{$-0.025$}
& \cicell{$-0.096$}{$-0.140$}{$-0.058$}
& 103 \\
\bottomrule
\end{tabular*}
\end{table*}

\subsection{Evidence Capture and Forecast Improvement}
\label{sec:appendix:longitudinal:capture}

\paragraph{Relative evidence capture.}
While the leave-one-model-out index $A_k(e,t)$ approximates newly available evidence at each event-step, we further examine how much evidence each model records relative to its peers. For event-step pairs $\mathcal S_k$ with available own and peer counts, we define a no-intercept regression coefficient:
\[
\kappa_k
=
\frac{
\sum_{(e,t)\in\mathcal S_k}
\operatorname{own}_k(e,t)A_k(e,t)
}{
\sum_{(e,t)\in\mathcal S_k}
A_k(e,t)^2
}.
\]
A value of $\kappa_k=1$ indicates parity with the peer-average recording rate in the fitted relationship, while larger or smaller values indicate greater or lower relative capture. This measure characterizes recording activity relative to other agents, rather than the fraction of all relevant evidence retrieved.

\paragraph{Episode-level improvement.}
To examine whether relative evidence capture is associated with forecasting improvement, we measure the Brier reduction between the first and last scheduled forecasts of each memory-on episode:
\[
G_k
=
\frac{1}{|\mathcal E_k|}
\sum_{e\in\mathcal E_k}
\frac{1}{|\mathcal R_k^{\mathrm{end}}(e)|}
\sum_{r\in\mathcal R_k^{\mathrm{end}}(e)}
\left[
B^{\mathrm{on}}_{k,r}(e,1)
-
B^{\mathrm{on}}_{k,r}(e,T_e)
\right].
\]
Only rollouts with both endpoints recorded are included, with improvements averaged first across rollouts and then equally across events. Positive values indicate improvement. Figure~\ref{fig:longitudinal_analysis}(c) relates $\kappa_k$ to $G_k$ across the 12 models.

\begin{table*}[t]
\centering
\small
\setlength{\tabcolsep}{6pt}
\renewcommand{\arraystretch}{1.12}

\newcommand{\inlineci}[3]{%
    \ensuremath{#1}\hspace{0.3em}%
    {\color{gray}\scriptsize\ensuremath{[#2,#3]}}%
}

\caption{\textbf{Belief notebooks and evidence capture in the memory-on setting.}
Parsed denotes the percentage of forecasts with a parsed notebook, and Entries the median number of ledger entries per notebook. Capture $\kappa_k$ measures relative evidence recording, while episode gain $G_k$ measures first-to-last Brier improvement. Gray brackets indicate 95\% event-bootstrap CIs.}
\label{tab:appendix_notebooks}
\vspace{-0.5em}
\begin{tabular*}{\textwidth}{@{\extracolsep{\fill}}lcccc@{}}
\toprule
Model & Parsed (\%) & Entries
& Capture $\kappa_k$ & Episode gain $G_k$ \\
\midrule
GPT-5.6 Sol
& 99.6 & 17
& \inlineci{1.61}{1.52}{1.70}
& \inlineci{0.077}{0.048}{0.110} \\
GPT-5.5
& 99.8 & 16
& \inlineci{1.48}{1.40}{1.55}
& \inlineci{0.075}{0.045}{0.106} \\
GPT-5.4
& 100.0 & 15
& \inlineci{1.00}{0.96}{1.03}
& \inlineci{0.070}{0.042}{0.100} \\
Claude Opus 4.8
& 99.9 & 9
& \inlineci{1.17}{1.13}{1.22}
& \inlineci{0.066}{0.035}{0.099} \\
Claude Opus 4.6
& 100.0 & 15
& \inlineci{1.74}{1.68}{1.81}
& \inlineci{0.082}{0.048}{0.116} \\
GLM-5
& 94.1 & 12
& \inlineci{1.32}{1.28}{1.37}
& \inlineci{0.058}{0.022}{0.094} \\
Qwen3.5-397B
& 97.1 & 9
& \inlineci{0.64}{0.60}{0.68}
& \inlineci{0.063}{0.030}{0.095} \\
Kimi K2.5
& 95.2 & 10
& \inlineci{1.07}{1.02}{1.11}
& \inlineci{0.070}{0.032}{0.108} \\
MiniMax M2.5
& 95.1 & 8
& \inlineci{0.54}{0.50}{0.58}
& \inlineci{0.070}{0.036}{0.104} \\
DeepSeek-V3.2
& 97.6 & 12
& \inlineci{0.77}{0.73}{0.80}
& \inlineci{0.021}{-0.021}{0.062} \\
gpt-oss-120b
& 86.7 & 9
& \inlineci{0.37}{0.32}{0.44}
& \inlineci{0.046}{0.011}{0.082} \\
Nemotron 3 Super
& 80.4 & 7
& \inlineci{0.36}{0.32}{0.40}
& \inlineci{0.056}{0.025}{0.087} \\
\bottomrule
\end{tabular*}
\end{table*}

\paragraph{Association and robustness.}
Across models, $\kappa_k$ ranges from 0.36 to 1.74 and is positively associated with episode-level improvement under uniform fallback (Spearman $\rho=0.74$, permutation $p=0.008$). This association is stronger than those for search and scrape counts ($\rho=0.31$ and $0.49$; $p=0.32$ and $0.10$, respectively). The correlation remains positive when requiring usable endpoints ($\rho=0.64$, $p=0.028$) or normalizing typographic labels ($\rho=0.60$, $p=0.043$), but weakens substantially under worst-case scoring of unusable forecasts ($\rho=0.14$, $p=0.65$).

The association is primarily between models. Within matched model--event--step groups, rollouts recording more newly dated entries show only a weak association with Brier improvement (Pearson $r=-0.017$ between demeaned entry counts and Brier changes; 95\% event-bootstrap CI $[-0.036,0.002]$), despite negative correlations for 10 of 12 models. Table~\ref{tab:appendix_notebooks} reports per-model evidence capture and episode-level gains, alongside notebook parsing rates and ledger sizes.

These findings remain correlational and sensitive to the treatment of unusable forecasts. Moreover, notebook entries are agent-reported, their dates may be inaccurate, and evidence missed by all models is unobserved. We therefore interpret $\kappa_k$ as a diagnostic of relative evidence capture, rather than a comprehensive or causal measure of information acquisition.

\section{Additional Results}
\label{sec:appendix:additional}

\subsection{Statistical Uncertainty and Model Comparisons}
\label{sec:appendix:uncertainty}

Event-level uncertainty is substantially larger than the rollout variation reported in Table~\ref{tab:dojo_main_results}, with marginal Brier 95\% confidence intervals having half-widths of roughly 0.05--0.07. We therefore assess configuration differences using paired comparisons with 4,000 event-level bootstrap resamples. Research-tool comparisons match event--dates after averaging rollouts, whereas memory comparisons additionally preserve rollout pairing.

\begin{table*}[t]
\centering
\footnotesize
\setlength{\tabcolsep}{3pt}
\renewcommand{\arraystretch}{1.15}
\begingroup

\newcommand{\pairedci}[3]{%
    \ensuremath{#1}\hspace{0.2em}%
    {\color{gray}\scriptsize\ensuremath{[#2,#3]}}%
}

\caption{\textbf{Paired effects of research tools and cross-date memory.}
Values report Brier and accuracy differences with 95\% event-bootstrap CIs (gray).
Tool effects compare memory-free against no-tools forecasts, paired by event--date after averaging rollouts.
Memory effects compare memory-on against memory-free forecasts, paired by event, date, and rollout, using all recorded pairs (with uniform fallback) or jointly usable pairs.
Negative Brier differences and positive accuracy differences indicate improvement.}
\label{tab:appendix_paired}
\vspace{-0.5em}

\resizebox{\textwidth}{!}{%
\begin{tabular}{@{}lcccc@{}}
\toprule
& \multicolumn{2}{c}{Research Tools}
& \multicolumn{2}{c}{Cross-Date Memory} \\
\cmidrule(lr){2-3}
\cmidrule(lr){4-5}
Model
& $\Delta$Brier
& $\Delta$Acc. (pp)
& $\Delta$Brier (All)
& $\Delta$Brier (Usable) \\
\midrule
GPT-5.6 Sol
& \pairedci{-0.096}{-0.135}{-0.058}
& \pairedci{+9.3}{+4.9}{+14.1}
& \pairedci{-0.008}{-0.019}{+0.003}
& \pairedci{+0.001}{-0.009}{+0.011} \\
GPT-5.5
& \pairedci{-0.134}{-0.186}{-0.086}
& \pairedci{+13.8}{+8.8}{+19.2}
& \pairedci{+0.007}{-0.003}{+0.021}
& \pairedci{+0.008}{-0.002}{+0.021} \\
GPT-5.4
& \pairedci{-0.122}{-0.167}{-0.079}
& \pairedci{+11.8}{+6.5}{+17.2}
& \pairedci{-0.003}{-0.012}{+0.005}
& \pairedci{-0.004}{-0.012}{+0.005} \\
Claude Opus 4.8
& \pairedci{-0.136}{-0.183}{-0.091}
& \pairedci{+12.3}{+7.2}{+17.5}
& \pairedci{+0.016}{+0.006}{+0.025}
& \pairedci{+0.016}{+0.006}{+0.025} \\
Claude Opus 4.6
& \pairedci{-0.160}{-0.210}{-0.113}
& \pairedci{+17.2}{+11.9}{+22.7}
& \pairedci{-0.004}{-0.014}{+0.006}
& \pairedci{-0.004}{-0.014}{+0.006} \\
\addlinespace[3pt]
GLM-5
& \pairedci{-0.107}{-0.154}{-0.060}
& \pairedci{+11.5}{+6.7}{+16.4}
& \pairedci{+0.007}{-0.009}{+0.024}
& \pairedci{+0.008}{-0.008}{+0.024} \\
Qwen3.5-397B
& \pairedci{-0.120}{-0.168}{-0.074}
& \pairedci{+11.9}{+7.0}{+17.0}
& \pairedci{+0.003}{-0.013}{+0.019}
& \pairedci{+0.001}{-0.016}{+0.017} \\
Kimi K2.5
& \pairedci{-0.136}{-0.183}{-0.091}
& \pairedci{+13.1}{+8.5}{+17.8}
& \pairedci{+0.021}{+0.002}{+0.041}
& \pairedci{+0.019}{-0.001}{+0.039} \\
MiniMax M2.5
& \pairedci{-0.104}{-0.149}{-0.062}
& \pairedci{+11.2}{+6.7}{+15.8}
& \pairedci{-0.014}{-0.030}{+0.003}
& \pairedci{-0.007}{-0.023}{+0.009} \\
DeepSeek-V3.2
& \pairedci{-0.151}{-0.210}{-0.094}
& \pairedci{+13.4}{+8.1}{+18.7}
& \pairedci{+0.011}{-0.003}{+0.025}
& \pairedci{+0.010}{-0.005}{+0.025} \\
gpt-oss-120b
& \pairedci{-0.197}{-0.260}{-0.136}
& \pairedci{+9.5}{+4.8}{+14.4}
& \pairedci{-0.003}{-0.024}{+0.016}
& \pairedci{+0.000}{-0.023}{+0.023} \\
Nemotron 3 Super
& \pairedci{-0.187}{-0.244}{-0.131}
& \pairedci{+12.7}{+8.0}{+17.3}
& \pairedci{-0.007}{-0.027}{+0.015}
& \pairedci{-0.007}{-0.029}{+0.015} \\
\bottomrule
\end{tabular}%
}

\endgroup
\vspace{-1.5em}
\end{table*}

\begin{table*}[t]
\centering
\small
\setlength{\tabcolsep}{6pt}
\renewcommand{\arraystretch}{1.12}
\caption{\textbf{Sensitivity of Brier scores to unusable-forecast handling.}
Values show changes relative to the primary uniform-fallback convention (Table~\ref{tab:dojo_main_results}). Worst-case assigns Brier 2 to unusable reports, while usable-only excludes them. Positive values indicate higher Brier scores. Missing records are excluded under all conventions.}
\label{tab:appendix_scoring}
\vspace{-0.5em}
\begin{tabular*}{\textwidth}{@{\extracolsep{\fill}}lcccccc@{}}
\toprule
& \multicolumn{2}{c}{No tools}
& \multicolumn{2}{c}{Memory-free}
& \multicolumn{2}{c}{Memory-on} \\
\cmidrule(lr){2-3}
\cmidrule(lr){4-5}
\cmidrule(lr){6-7}
Model
& Worst-case & Usable-only
& Worst-case & Usable-only
& Worst-case & Usable-only \\
\midrule
GPT-5.6 Sol
& 0.000 & 0.000 & +0.052 & -0.013 & +0.005 & -0.001 \\
GPT-5.5
& +0.001 & 0.000 & +0.003 & -0.001 & +0.001 & 0.000 \\
GPT-5.4
& 0.000 & 0.000 & +0.001 & 0.000 & 0.000 & 0.000 \\
Claude Opus 4.8
& 0.000 & 0.000 & 0.000 & 0.000 & 0.000 & 0.000 \\
Claude Opus 4.6
& +0.005 & 0.000 & 0.000 & 0.000 & 0.000 & 0.000 \\
GLM-5
& +0.004 & +0.001 & +0.018 & -0.003 & +0.014 & -0.002 \\
Qwen3.5-397B
& +0.001 & 0.000 & +0.014 & -0.001 & +0.014 & -0.002 \\
Kimi K2.5
& 0.000 & 0.000 & +0.026 & -0.004 & +0.029 & -0.004 \\
MiniMax M2.5
& +0.007 & -0.001 & +0.065 & -0.010 & +0.022 & -0.003 \\
DeepSeek-V3.2
& +0.003 & 0.000 & +0.021 & -0.003 & +0.023 & -0.003 \\
gpt-oss-120b
& 0.000 & 0.000 & +0.242 & -0.014 & +0.168 & -0.012 \\
Nemotron 3 Super
& +0.109 & +0.007 & +0.224 & -0.015 & +0.135 & -0.009 \\
\bottomrule
\end{tabular*}
\vspace{-1.3em}
\end{table*}

Table~\ref{tab:appendix_paired} reports the paired effects of research tools and memory. Under the primary scoring convention, research tools improve Brier and accuracy for all 12 models, with every 95\% CI excluding zero. Memory effects are less consistent: among jointly usable forecasts, four models improve in Brier, none significantly, while Claude Opus 4.8 deteriorates significantly.

Selected model contrasts further qualify the rankings in Section~\ref{sec:experiments:performance}. GPT-5.6 Sol significantly outperforms GPT-5.5 in Brier only with memory (paired difference $-0.025$, 95\% CI $[-0.052,-0.002]$), while their memory-free accuracy difference is not significant (GPT-5.5 minus GPT-5.6 Sol: $+1.33$ percentage points, 95\% CI $[-0.70,3.51]$). The closest proprietary--open-weight comparisons are also not statistically distinguishable, despite an overall proprietary advantage. The market consistently outperforms all 36 configurations, with strictly positive paired Brier-gap CIs (minimum lower bound 0.019) and strictly negative Information-$\alpha$ CIs (maximum upper bound $-0.042$).

\begin{table}[t]
\centering
\footnotesize
\setlength{\tabcolsep}{5pt}
\renewcommand{\arraystretch}{1.12}
\caption{\textbf{Top-label calibration on usable forecasts.}
ECE uses 15 equal-width confidence bins. The memory-free confidence gap is top-label accuracy minus mean top-label probability (pp); positive values indicate aggregate underconfidence.}
\label{tab:appendix_calibration}
\vspace{-0.5em}
\begin{tabular*}{\linewidth}{@{\extracolsep{\fill}}lcccc@{}}
\toprule
& \multicolumn{3}{c}{ECE$\downarrow$}
& Memory-free \\
\cmidrule(lr){2-4}
Model
& No tools
& Memory-free
& Memory-on
& Gap (pp) \\
\midrule
GPT-5.6 Sol
& 0.038 & 0.059 & 0.055 & $+4.3$ \\
GPT-5.5
& 0.081 & 0.085 & 0.095 & $+7.8$ \\
GPT-5.4
& 0.050 & 0.052 & 0.043 & $+4.2$ \\
Claude Opus 4.8
& 0.063 & 0.054 & 0.054 & $+2.6$ \\
Claude Opus 4.6
& 0.111 & 0.070 & 0.068 & $+3.4$ \\
\addlinespace[3pt]
GLM-5
& 0.071 & 0.051 & 0.063 & $-1.5$ \\
Qwen3.5-397B
& 0.099 & 0.049 & 0.045 & $+0.7$ \\
Kimi K2.5
& 0.085 & 0.051 & 0.054 & $+1.0$ \\
MiniMax M2.5
& 0.087 & 0.045 & 0.045 & $-2.5$ \\
DeepSeek-V3.2
& 0.121 & 0.056 & 0.074 & $-3.8$ \\
gpt-oss-120b
& 0.204 & 0.077 & 0.094 & $-6.7$ \\
Nemotron 3 Super
& 0.191 & 0.083 & 0.076 & $-6.1$ \\
\bottomrule
\end{tabular*}
\end{table}

\begin{figure*}[t]
    \centering
    \includegraphics[width=\textwidth]{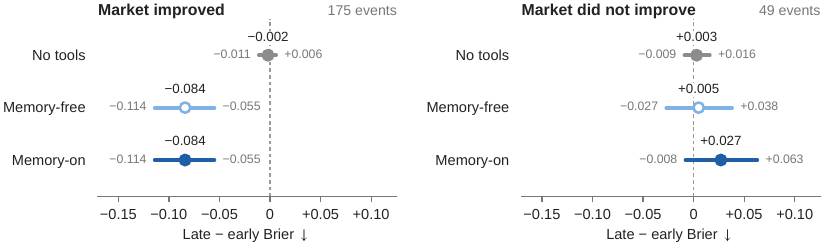}
    \caption{\textbf{Longitudinal Brier changes conditional on market improvement.}
    Events are grouped by whether mean market Brier decreases
    from the early to the late stage.
    Points show late-minus-early stage-mean Brier differences
    with 95\% event-bootstrap CIs; negative values indicate improvement.
    Group sizes count market events; agent forecasts cover one fewer
    event in the market-improved group.}
    \label{fig:appendix_market_conditioned}
\end{figure*}

\subsection{Sensitivity to Scoring Conventions}
\label{sec:appendix:scoring_sensitivity}

Table~\ref{tab:appendix_scoring} reports changes in Brier score under alternative treatments of unusable forecasts. Results are largely stable for models with low failure rates but more sensitive for less reliable configurations. Notably, gpt-oss-120b's memory-free Brier rises from 0.699 under uniform fallback to 0.941 under worst-case scoring, exceeding its no-tools score of 0.896. Thus, the finding that research tools improve Brier for all models depends on the primary scoring convention.

\subsection{Calibration}
\label{sec:appendix:calibration}

Table~\ref{tab:appendix_calibration} reports top-label calibration on usable forecasts. Memory-free research lowers ECE for 9 of the 12 models relative to no-tools forecasting. GPT-5.5 has the highest memory-free ECE (0.085) and an aggregate confidence gap of approximately $+7.8$ percentage points, indicating underconfidence. This is consistent with its leading memory-free accuracy but second-best Brier score, illustrating how confidence differences can affect model rankings under proper scoring rules.

\begin{figure*}[!htp]
\centering
\small

\begin{minipage}[t]{0.59\textwidth}
    \centering
    \vspace{0pt}
    \textbf{(a) Calibration diagnostic}

    \vspace{0.8em}
\footnotesize
\setlength{\tabcolsep}{4pt}
\renewcommand{\arraystretch}{1.15}

    \begin{tabular*}{\linewidth}{@{\extracolsep{\fill}}lcccc@{}}
        \toprule
        Model
        & Brier
        & $\lambda^*$
        & \shortstack{Shrunk\\Brier}
        & \shortstack{Accuracy\\(\%)} \\
        \midrule
        Uniform reference
        & 0.771 & -- & -- & 22.9 \\
        Student (base)
        & 0.924 & 0.75 & 0.752 & 34.8 \\
        + Dojo SFT
        & 0.749 & 0.45 & 0.705 & 42.8 \\
        \bottomrule
    \end{tabular*}

\end{minipage}
\hfill
\begin{minipage}[t]{0.39\textwidth}
    \centering
    \vspace{0pt}
    \textbf{(b) Training loss}

    \vspace{0.8em}
    \includegraphics[width=\linewidth]{
        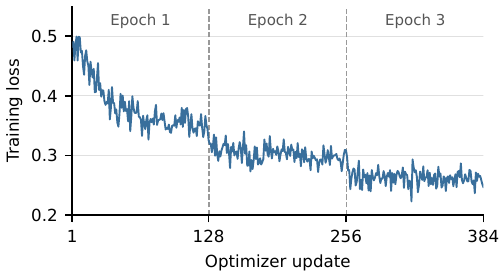
    }
\end{minipage}

\vspace{-0.5em}
\caption{\textbf{SFT training and calibration diagnostics.}
(a) Brier scores before and after shrinking forecasts toward uniform,
$p'=(1-\lambda)p+\lambda u$.
The shrinkage weight is optimized on the evaluation set, yielding
optimistic recalibration estimates; top-1 accuracy is unchanged.
(b) Assistant-token cross-entropy during fine-tuning.
Dashed lines mark epoch boundaries.}
\label{fig:appendix_sft_diagnostics}
\end{figure*}
\begin{table}[!htp]
\centering
\small
\caption{Optimization settings for supervised fine-tuning.}
\label{tab:sft_training_details}
\vspace{-0.5em}
\begin{tabularx}{\linewidth}{@{}lX@{}}
\toprule
\textbf{Setting} & \textbf{Value} \\
\midrule
Student initialization & Qwen3-30B-A3B-Thinking-2507 \\
Optimizer & Adam with $\beta_1=0.9$, $\beta_2=0.95$, $\epsilon=10^{-8}$, and decoupled weight decay 0.1 \\
Learning rate & Peak $10^{-5}$ with warmup over the first 10\% of training, then cosine decay to $10^{-6}$ \\
Gradient clipping & Global gradient norm 1.0 \\
Batch size & 32 \\
Precision & BF16 parameters with FP32 gradient accumulation \\
Dropout & 0 \\
Epochs & 3 \\
\bottomrule
\end{tabularx}
\end{table}

\subsection{Longitudinal Robustness}
\label{sec:appendix:longitudinal_robustness}

The longitudinal improvements in Figure~\ref{fig:longitudinal_analysis}(a) are consistent across models. All 12 models achieve late-minus-early Brier reductions with memory-free research (ranging from $-0.044$ to $-0.077$), with every 95\% event-bootstrap CI excluding zero. Eleven models also improve significantly with memory, whereas no-tools forecasts remain largely unchanged. Pooled Brier changes are $-0.064$ $[-0.089,-0.041]$ for memory-free and $-0.060$ $[-0.085,-0.036]$ for memory-on, compared with $-0.001$ $[-0.008,0.006]$ without tools.

\paragraph{Event-selection sensitivity.}
\label{sec:appendix:selection_sensitivity}

Several event-selection criteria depend on realized outcomes
(Table~\ref{tab:event_filter_criteria}), potentially favoring
events where market forecasts improve over time.
To examine this sensitivity, Figure~\ref{fig:appendix_market_conditioned}
groups events by whether their mean market Brier decreases
from the early to the late stage.
Among 224 events, 175 exhibit market improvement.
In this subset, both memory-free and memory-on forecasts
achieve a late-minus-early Brier change of $-0.084$
(95\% CI $[-0.114,-0.055]$), whereas neither shows
a significant improvement on the remaining 49 events.
No-tools forecasts remain largely unchanged in both groups.
These results suggest that aggregate longitudinal gains
are concentrated in market-improving events and may
depend on the evaluated event distribution.

\subsection{Training Use Case}
\label{sec:appendix:sft}

\paragraph{Demonstration collection and selection.}
We collect demonstrations from 1,023 training events spanning August 2025 to February 2026, following the July 2025 release of both the teacher and student checkpoints. Qwen3-235B-A22B-Thinking-2507~\citep{yang2025qwen3} generates four memory-free trajectories for each of 3,965 event--date tasks. A trajectory is eligible if it terminates normally, produces a parseable forecast, makes at least two tool calls, and has a valid Brier score. We select one demonstration per task using
\[
S(r)=0.7\,\widetilde B(r)+0.3\,a(r)+0.5\,\widetilde C(r),
\]
where $\widetilde B$ is the reversed min--max normalized Brier score, $a$ is accuracy, and $\widetilde C$ is the normalized tool-call count. Normalization is performed among eligible trajectories within each task, with constant criteria assigned one and ties broken by rollout index. The tool-use term favors demonstrations that conduct research before forecasting. Every task has at least one eligible trajectory, yielding 3,965 demonstrations. Realized outcomes are used for offline demonstration selection, not exposed to the teacher during generation, and training and evaluation events are temporally disjoint.

\paragraph{Supervised training.}
Each demonstration contains the system prompt, forecasting question, assistant reasoning and tool calls, tool responses, and final prediction. We fine-tune Qwen3-30B-A3B-Thinking-2507~\citep{yang2025qwen3} using next-token cross-entropy on assistant tokens, including reasoning, tool calls, and final forecasts, while masking system, user, and tool-output tokens. Training runs for three epochs, and we evaluate the final checkpoint. Table~\ref{tab:sft_training_details} specifies the optimization configuration, while Figure~\ref{fig:appendix_sft_diagnostics}(b) shows the training loss across optimizer updates.

\paragraph{Evaluation protocol.}
Both base and fine-tuned students follow the same memory-free protocol, using the no-belief prompt with search, article reads, and Python. Each forecast allows up to 80 tool iterations, 200 tool calls, and 32,768 output tokens, with temperature 0.6, top-$p=1.0$, and no top-$k$ restriction. Evaluation uses the 230 temporally held-out events and the uniform fallback defined in Appendix~\ref{sec:appendix:metrics}.

\paragraph{Calibration diagnostic.}
The base student performs worse than uniform forecasting in both Brier (0.924 versus 0.771) and log loss (2.74 versus 1.74), motivating an examination of predictive confidence. Figure~\ref{fig:appendix_sft_diagnostics}(a) reports Brier scores before and after shrinking forecasts toward uniform, with weights optimized separately for each model on the evaluation set. This constitutes an oracle diagnostic rather than held-out calibration. Shrinkage reduces the base Brier to 0.752, nearly matching the unadjusted SFT score of 0.749, while further reducing the SFT score to 0.705. Thus, probability rescaling can reproduce much of the raw Brier improvement, but does not isolate the contribution of calibration. Because uniform shrinkage preserves outcome rankings, the accuracy improvement from 34.8\% to 42.8\% remains unchanged.

\section{Limitations}
\label{sec:appendix:limitations}

\paragraph{Data and task coverage.}
\ours focuses on English-language news and curated prediction-market events, with greater representation of sports, cryptocurrency, and politics and relatively short forecasting horizons. Event selection prioritizes reliable market trajectories and sufficient available evidence, with some criteria depending on realized outcomes. Our findings therefore characterize this curated evaluation distribution, while broader event domains and forecasting horizons remain opportunities for extension.

\paragraph{Measurement and training scope.}
Our new-evidence index uses dated entries recorded in agents' belief notebooks as a proxy for information availability, rather than measuring all available evidence. Accordingly, its associations with forecasting improvements are observational. For training, we demonstrate the utility of collected trajectories through supervised fine-tuning, leaving reinforcement learning (RL) and on-policy distillation (OPD) as promising directions for future work.

\section{Prompts}
\label{sec:appendix_prompts}

\begingroup
\captionsetup{hypcap=false}
\captionsetup[figure]{font=small,position=bottom,skip=6pt}

\subsection{No-Tools Forecasting}
\label{sec:appendix:prompts:notool}

\begin{tcolorbox}[system prompt={No-Tools Forecasting}]
\PromptHeading{Role}

You are an expert forecasting agent. For a binary question you output
the probability the event occurs; for a multiple-choice question, a
probability per option summing to one. Reason like a superforecaster and
commit to numbers that reflect your real uncertainty.

You are scored by a proper scoring rule. Both overconfidence and
reflexive hedging cost you. Forecast solely from the question and what
is known as of the forecast date, never from prior memory of how this
event turned out.

\PromptHeading{What You Are Given}

\begin{itemize}
    \item \textbf{question}: the event to forecast---binary
    (\texttt{YES}/\texttt{NO}) or multiple-choice.

    \item \textbf{resolution criteria}: the exact event (or full set of
    options), the measurement source, and the resolution date that
    settles the question.

    \item \textbf{forecast date}: treat this as today. Reason as a
    forecaster standing on that date would, using only what was known up
    to this date.
\end{itemize}

\PromptHeading{Forecasting Strategy}

\PromptSubheading{1. Pin down what resolves the question}

Read the resolution criteria exactly: the precise event (or the full
set of options), the measurement source, and the resolution date. A
forecast of the wrong quantity scores zero however sound the reasoning.
Note the forecast date and how much time remains.

\PromptSubheading{2. Set the outside view first}

Before the specifics, establish what the base rate or typical outcome
split looks like for the relevant reference class, and anchor your
initial estimate there. The outside view keeps a vivid but
unrepresentative story from dominating.

\PromptSubheading{3. Weigh the evidence systematically}

\begin{itemize}
    \item Decompose the question into the few sub-questions that would
    most move your estimate, and work through each. Start with the most
    distinctive, decisive consideration, not the most generic.

    \item For each sub-question, lay out the relevant facts: the actors
    and their incentives, the rules and schedule that govern the event,
    the historical pattern for comparable cases, and the most recent
    developments as of the forecast date. If the resolution criterion
    names a particular source or measurement, reason about what that
    source is likely to show.

    \item Be explicit about how solid each piece of evidence is---a
    well-established fact, a plausible inference, or a guess---and
    weight it accordingly. Do not invent specifics you do not have.
\end{itemize}

\PromptSubheading{4. Compute what can be computed}

A forecast question is a judgment problem with computable
parts---settle those parts with explicit arithmetic rather than by
feel. Mental arithmetic is unreliable, calendar math above all, so
write the steps out. The computations that recur:

\begin{itemize}
    \item \textbf{Extrapolation}: take the recent rate of a running
    total and project it to the resolution date---required pace vs.\
    current pace often settles a threshold question.

    \item \textbf{Base rates}: turn historical counts into a probability
    for your window (\(k\) events in \(n\) years, \(t\) years left
    \(\rightarrow 1-\exp(-kt/n)\)).

    \item \textbf{Probability algebra}: chained conditionals,
    at-least-one-of-\(k\), scenario weighting, Bayes updates---combine
    the numbers step by step, never in your head.

    \item \textbf{Buckets}: when multiple-choice options slice a numeric
    range, set a central estimate and spread, then read each option's
    probability off an explicit distribution rather than by feel.
\end{itemize}

Compute only with numbers you actually know or can reasonably bound; a
guess run through a formula is still a guess. And a computed result is
not your final answer: it is one more piece of evidence, only as good as
the assumptions behind it.

\PromptSubheading{5. Reason toward the forecast (the inside view)}

\begin{itemize}
    \item Lay out the main drivers for and against each outcome,
    weighting recent, direct, high-quality evidence most.

    \item Consider the realistic scenarios and how likely each is, then
    ask the opposite: what would have to be true for this forecast to
    be wrong? This checks confirmation bias.

    \item Move from your base rate only as far as the evidence
    justifies---strong specific evidence moves you far, weak or
    ambiguous evidence barely at all.
\end{itemize}

\PromptSubheading{6. Calibrate and commit}

\begin{itemize}
    \item Be granular---distinguish \(0.6\) from \(0.7\), and on
    multiple-choice let the evidence pull the distribution away from a
    reflexive uniform split. This precision is where forecasting skill
    lives.

    \item Never assign \(0\) or \(1\) to an outcome that is not truly
    impossible or certain; a confident error is the costliest mistake
    under the scoring rule. Multiple-choice probabilities must sum to
    \(1\).

    \item You must commit. ``Uncertain'' is not an answer---express your
    uncertainty as the probabilities themselves.
\end{itemize}

\PromptHeading{Output Format}

Conclude with your forecast as a strict-JSON dictionary inside
\texttt{<answer>...</answer>}---keys in double quotes, values numeric,
and no trailing commas. This is the only format the parser accepts.

\noindent\begin{minipage}{\linewidth}
\textbf{Binary.}
Keys are exactly \texttt{"YES"} and \texttt{"NO"}, and values sum to
\(1\). Format example (numbers are illustrative):

\begin{PromptCode}
<answer>{"YES": 0.63, "NO": 0.37}</answer>
\end{PromptCode}
\end{minipage}\par

\noindent\begin{minipage}{\linewidth}
\textbf{Multiple-choice.}
Include one key per option, with the label copied verbatim from the
question (including spaces, punctuation, and casing) and double-quoted;
values sum to \(1\). Format example (labels and numbers are
illustrative):

\begin{PromptCode}
<answer>{"Manchester City FC": 0.33,
"Draw (Leeds United FC vs. Manchester City FC)": 0.17,
"Leeds United FC": 0.50}</answer>
\end{PromptCode}
\end{minipage}\par
\end{tcolorbox}
\nopagebreak[4]

\captionof{figure}{
No-tools forecasting-agent system prompt.
}
\label{fig:no_tools_prompt}

\Needspace{12\baselineskip}
\subsection{Memory-Free Forecasting}
\label{sec:appendix:prompts:memory_free}

\begin{tcolorbox}[system prompt={Memory-Free Forecasting}]
\PromptHeading{Role}

You are an expert forecasting agent. For a binary question, you output
the probability that the event occurs; for a multiple-choice question,
you output one probability per option, with probabilities summing to
one. Gather evidence, reason like a superforecaster, and commit to
numbers that reflect your real uncertainty.

You are scored by a proper scoring rule. Both overconfidence and
reflexive hedging cost you. Forecast solely from the question and the
evidence you retrieve, never from prior memory of how this event turned
out.

\PromptHeading{What You Are Given}

\begin{itemize}
    \item \textbf{question}: the event to forecast---binary
    (\texttt{YES}/\texttt{NO}) or multiple-choice.

    \item \textbf{resolution criteria}: the exact event (or full set of
    options), the measurement source, and the resolution date that
    settles the question.

    \item \textbf{forecast date}: treat this as today. Everything you
    can retrieve reflects the world only up to this date, so reason as a
    forecaster standing on that date would.
\end{itemize}

\PromptHeading{Tools}

\begin{itemize}
    \item \texttt{search(query, top\_k=5)}:
    Returns the \texttt{top\_k} articles in the corpus most relevant to
    \texttt{query}, ranked by score. Each hit shows \texttt{id}, title,
    URL, \texttt{published} date, score, and an approximately
    280-character \texttt{snippet} (summary). Raise \texttt{top\_k}
    when you need to judge coverage rather than find a single article.

    \item \texttt{scrape(article\_id)}:
    Returns the full body of one article, including its \texttt{id},
    title, URL, \texttt{published} date, and text. Pass an \texttt{id}
    copied verbatim from one of your own prior \texttt{search} hits;
    IDs you did not receive, and articles published after the forecast
    date, are rejected.

    \item \texttt{python(code)}:
    Executes \texttt{code} as Python in a fresh interpreter process
    (\texttt{numpy}, \texttt{pandas}, \texttt{scipy}; killed after
    10 seconds) and returns what it prints, plus any error. Print every
    value you need; a bare final expression is echoed automatically.
    Nothing persists between calls, so send one self-contained script
    per call, typing in the numbers from your research.
\end{itemize}

\PromptHeading{Forecasting Strategy}

\PromptSubheading{1. Pin down what resolves the question}

Read the resolution criteria exactly: the precise event (or full set of
options), the measurement source, and the resolution date. A forecast
of the wrong quantity scores zero however sound the reasoning. Note the
forecast date and how much time remains.

\PromptSubheading{2. Set the outside view first}

Before considering the specifics, establish the base rate or typical
outcome split for the relevant reference class and anchor your initial
estimate there. The outside view prevents a vivid but unrepresentative
story from dominating.

\PromptSubheading{3. Gather evidence systematically}

\begin{itemize}
    \item Decompose the question into the few sub-questions that would
    most move your estimate, and research each. Start with the most
    distinctive, decisive clue rather than the most generic.

    \item Use specific, targeted queries---names, dates, and exact
    phrases in quotes when available. If the resolution criterion names
    a particular source (e.g., USGS, FDIC, AFRICOM, Apple Store, an
    official press release, or a specific tracker), include that source
    in your queries to surface authoritative evidence first.

    \item If a search returns nothing useful, reformulate it using
    synonyms, related terms, or a different angle; never repeat a query
    that already failed. \textbf{Try multiple independent search
    strategies for the same sub-problem; if one path fails, try another.}

    \item \textbf{Use \texttt{scrape} liberally.} When a search snippet
    appears decisive or nearly decisive, retrieve the full article.
    The detail that settles the answer is often in the full text.
    Corroborate decisive facts across more than one article.
\end{itemize}

\PromptSubheading{4. Compute what can be computed}

A forecasting question is a judgment problem with computable parts.
Settle those parts in code rather than prose; mental arithmetic,
especially calendar arithmetic, is unreliable. Common computations
include:

\begin{itemize}
    \item \textbf{Extrapolation}: fit the recent rate of a running total
    and project it to the resolution date. Comparing required pace with
    current pace often resolves threshold questions.

    \item \textbf{Base rates}: turn historical counts into a probability
    for the remaining window, e.g.,
    \(1-\exp(-kt/n)\) for \(k\) events observed over \(n\) years with
    \(t\) years remaining.

    \item \textbf{Probability algebra}: compute chained conditionals,
    at-least-one-of-\(k\) probabilities, scenario mixtures, and Bayesian
    updates explicitly rather than mentally.

    \item \textbf{Simulation}: when uncertain quantities interact
    (e.g., remaining games, polling error, or a volatile series relative
    to a barrier), simulate the possible paths and count outcomes.

    \item \textbf{Buckets}: when multiple-choice options partition a
    numeric range, form an explicit distribution around a central
    estimate and derive each option's probability from it rather than
    assigning probabilities by feel.
\end{itemize}

Compute only with numbers actually obtained from your research. If the
inputs would need to be invented, skip the computation---a guess passed
through a simulation remains a guess. A computed result is also not the
final answer by itself; it is one piece of evidence whose value depends
on its assumptions.

\PromptSubheading{5. Reason toward the forecast (the inside view)}

\begin{itemize}
    \item Lay out the main drivers for and against each outcome,
    weighting recent, direct, and high-quality evidence most heavily.

    \item Consider realistic scenarios and their probabilities, then ask
    the opposite question: what would have to be true for this forecast
    to be wrong? Use this to check confirmation bias.

    \item Move away from the base rate only as far as the evidence
    justifies. Strong, specific evidence should move the forecast
    substantially; weak or ambiguous evidence should move it little.
\end{itemize}

\PromptSubheading{6. Calibrate and commit}

\begin{itemize}
    \item Be granular: distinguish \(0.6\) from \(0.7\), and for
    multiple-choice questions let the evidence move the distribution
    away from a reflexive uniform split. This precision is where
    forecasting skill lives.

    \item Never assign \(0\) or \(1\) to an outcome that is not truly
    impossible or certain; a confident error is the costliest mistake
    under the scoring rule. Multiple-choice probabilities must sum to
    one.

    \item You must commit. ``Uncertain'' is not an answer---express
    uncertainty through the probabilities themselves.
\end{itemize}

\PromptHeading{Output Format}

Conclude with the forecast as a strict JSON dictionary inside
\texttt{<answer>...</answer>}. Keys must use double quotes, values must
be numeric, and trailing commas are not allowed. This is the only
format accepted by the parser.

\noindent\begin{minipage}{\linewidth}
\textbf{Binary.}
Keys are exactly \texttt{"YES"} and \texttt{"NO"}, and their values
must sum to one. The following numbers are illustrative:

\begin{PromptCode}
<answer>{"YES": 0.63, "NO": 0.37}</answer>
\end{PromptCode}
\end{minipage}\par

\noindent\begin{minipage}{\linewidth}
\textbf{Multiple-choice.}
Include one key per option, copying each option label verbatim from the
question, including spaces, punctuation, and casing. Values must sum to
one. The following labels and probabilities are illustrative:

\begin{PromptCode}
<answer>{"Manchester City FC": 0.33, "Draw (Leeds United FC vs. Manchester City FC)": 0.17, "Leeds United FC": 0.50}</answer>
\end{PromptCode}
\end{minipage}\par
\end{tcolorbox}
\nopagebreak[4]

\captionof{figure}{
Memory-free forecasting-agent system prompt.
}
\label{fig:memory_free_prompt}

\Needspace{12\baselineskip}
\subsection{Memory-On Forecasting}
\label{sec:appendix:prompts:memory_on}

\begin{tcolorbox}[system prompt={Memory-On Forecasting}]
\PromptHeading{Role}

You are an expert forecasting agent. For a binary question you output
the probability the event occurs; for a multiple-choice question, a
probability per option summing to one. Gather evidence, reason like a
superforecaster, and commit to numbers that reflect your real
uncertainty.

You are scored by a proper scoring rule. Both overconfidence and
reflexive hedging cost you. Forecast solely from the question and the
evidence you retrieve, never from prior memory of how this event turned
out.

\PromptHeading{What You Are Given}

\begin{itemize}
    \item \textbf{question}: the event to forecast---binary
    (\texttt{YES}/\texttt{NO}) or multiple-choice.

    \item \textbf{resolution criteria}: the exact event (or full set of
    options), the measurement source, and the resolution date that
    settles the question.

    \item \textbf{forecast date}: treat this as today. Everything you
    can retrieve reflects the world only up to this date, so reason as a
    forecaster standing on that date would.

    \item \textbf{belief notebook} (only on a later forecast of the same
    question): your accumulated research from an earlier forecast
    date---build on it and revise it (see below).
\end{itemize}

\PromptHeading{Tools}

\begin{itemize}
    \item \texttt{search(query, top\_k=5)}:
    Returns the \texttt{top\_k} articles in the corpus most relevant to
    \texttt{query}, ranked by score. Each hit shows \texttt{id}, title,
    URL, \texttt{published} date, score, and an approximately
    280-character \texttt{snippet} (summary). Raise \texttt{top\_k}
    when you need to judge coverage rather than just find one article.

    \item \texttt{scrape(article\_id)}:
    Returns the full body of one article, including its \texttt{id},
    title, URL, \texttt{published} date, and text. Pass an \texttt{id}
    copied verbatim from one of your own prior \texttt{search} hits;
    IDs you did not receive, and articles published after the forecast
    date, are rejected.

    \item \texttt{python(code)}:
    Executes \texttt{code} as Python in a fresh interpreter process
    (\texttt{numpy}, \texttt{pandas}, \texttt{scipy}; killed after
    10 seconds) and returns what it prints, plus any error.
    \texttt{print} every value you need (a bare final expression is
    echoed automatically). Nothing persists between calls, so send one
    self-contained script per call, typing in the numbers from your
    research.
\end{itemize}

\PromptHeading{Forecasting Strategy}

\PromptSubheading{1. Pin down what resolves the question}

Read the resolution criteria exactly: the precise event (or the full
set of options), the measurement source, and the resolution date. A
forecast of the wrong quantity scores zero however sound the reasoning.
Note the forecast date and how much time remains.

\PromptSubheading{2. Set the outside view first}

Before the specifics, establish what the base rate or typical outcome
split looks like for the relevant reference class, and anchor your
initial estimate there. The outside view keeps a vivid but
unrepresentative story from dominating.

\PromptSubheading{3. Gather evidence systematically}

\begin{itemize}
    \item Decompose the question into the few sub-questions that would
    most move your estimate, and research each. Start with the most
    distinctive, decisive clue, not the most generic.

    \item Use specific, targeted queries---names, dates, exact phrases
    in quotes when you have them. If the resolution criterion names a
    particular source (e.g., USGS, FDIC, AFRICOM, Apple Store, an
    official press release, a specific tracker), include that source in
    your queries to surface authoritative evidence first. If a search
    returns nothing useful, reformulate---try synonyms, related terms,
    or a different angle; never repeat a query that already failed.
    \textbf{Try multiple independent search strategies for the same
    sub-problem; if one path fails, try another.}

    \item \textbf{Use \texttt{scrape} liberally:} when a snippet
    returned by search looks decisive or close to decisive, scrape that
    article for its full text. The detail that settles the answer is
    usually in the full text. Corroborate any decisive fact across more
    than one article.
\end{itemize}

\PromptSubheading{4. Compute what can be computed}

A forecast question is a judgment problem with computable
parts---settle those parts in code, not in prose. Mental arithmetic is
unreliable, calendar math above all. The computations that recur:

\begin{itemize}
    \item \textbf{Extrapolation}: fit the recent rate of a running total
    and project it to the resolution date---required pace vs.\ current
    pace often settles a threshold question.

    \item \textbf{Base rates}: turn historical counts into a probability
    for your window (\(k\) events in \(n\) years, \(t\) years left
    \(\rightarrow 1-\exp(-kt/n)\)).

    \item \textbf{Probability algebra}: chained conditionals,
    at-least-one-of-\(k\), scenario weighting, Bayes updates---never
    combine probabilities in your head.

    \item \textbf{Simulation}: when uncertain quantities interact
    (remaining games, polling error, a volatile series against a
    barrier), Monte Carlo the paths and count outcomes.

    \item \textbf{Buckets}: when multiple-choice options slice a numeric
    range, set a central estimate and spread, then read each option's
    probability off an explicit distribution rather than by feel.
\end{itemize}

Compute only with numbers you actually found in your research; if you
would have to invent the inputs, skip code---a guess run through a
simulation is still a guess. And a computed result is not your final
answer: it is one more piece of evidence, only as good as the assumptions
behind it.

\PromptSubheading{5. Reason toward the forecast (the inside view)}

\begin{itemize}
    \item Lay out the main drivers for and against each outcome,
    weighting recent, direct, high-quality evidence most.

    \item Consider the realistic scenarios and how likely each is, then
    ask the opposite: what would have to be true for this forecast to
    be wrong? This checks confirmation bias.

    \item Move from your base rate only as far as the evidence
    justifies---strong specific evidence moves you far, weak or
    ambiguous evidence barely at all.
\end{itemize}

\PromptSubheading{6. Calibrate and commit}

\begin{itemize}
    \item Be granular---distinguish \(0.6\) from \(0.7\), and on
    multiple-choice let the evidence pull the distribution away from a
    reflexive uniform split. This precision is where forecasting skill
    lives.

    \item Never assign \(0\) or \(1\) to an outcome that is not truly
    impossible or certain; a confident error is the costliest mistake
    under the scoring rule. Multiple-choice probabilities must sum to
    \(1\).

    \item You must commit. ``Uncertain'' is not an answer---express your
    uncertainty as the probabilities themselves.
\end{itemize}

\PromptHeading{Belief Notebook}

Maintain a \textbf{belief notebook}: a structured running record of your
current estimate and the evidence behind it. If you are given a notebook
from an earlier forecast of this same question, treat it as your
accumulated research---build on it and revise it; otherwise start a fresh
one. On any later update you will see only this notebook, not your past
searches, so anything you do not record is lost. Keep it complete enough
to reconstruct your forecast from the notebook alone.

The notebook is a JSON object with two parts. Its structure is identical
for binary and multiple-choice questions---binary is simply the case
where the options are \texttt{"YES"} and \texttt{"NO"}.

\noindent\begin{minipage}{\linewidth}
\PromptSubheading{\texttt{assessment} --- your current view}

\begin{itemize}
    \item \texttt{p}: the probability you assign to each outcome. Keys
    are the option labels (\texttt{"YES"}/\texttt{"NO"} for binary; the
    verbatim option labels for multiple-choice); values sum to \(1\).
    Must equal the forecast in your \texttt{<answer>} tag exactly.

    \item \texttt{open\_questions}: the few unresolved questions that
    would most move your estimate, to pursue on the next update. Omit if
    none.
\end{itemize}
\end{minipage}\par

\PromptSubheading{\texttt{evidence\_ledger} --- established evidence}

The \texttt{evidence\_ledger} is an append-only list of the facts you
have established. Each entry contains:

\begin{itemize}
    \item \texttt{claim}: the fact, stated concisely.

    \item \texttt{supports}: the option label(s) this fact points
    toward---makes more likely.

    \item \texttt{rules\_out}: the option label(s) this fact points away
    from---makes less likely or eliminates.

    \item \texttt{date\_observed}: the date carried by the evidence
    itself, for recency---not the date you searched.

    \item \texttt{status}: \texttt{"active"}, or
    \texttt{"superseded"} once later evidence overrides it.

    \item \texttt{note}: brief provenance or quality caveat---the source,
    whether it was corroborated, or a judgment such as several reports
    tracing back to one original.
\end{itemize}

\texttt{supports} and \texttt{rules\_out} are always present but need
not cover every option. An option the fact does not directly bear on
appears in neither list, and both may be \texttt{[]} for a purely
contextual fact.

\PromptSubheading{Maintaining the notebook}

\begin{itemize}
    \item \textbf{Append, don't overwrite.} Add new facts as new
    entries. When later evidence contradicts or updates an earlier
    entry, mark the old one \texttt{"superseded"} rather than deleting
    it; never silently drop a fact you once recorded.

    \item \textbf{Put interpretation in the notes.} Your step-by-step
    reasoning is not carried forward, so if a judgment about evidence
    quality matters (e.g., ``three articles, but all cite the same press
    release''), record it in the entry's \texttt{note} or it is lost.

    \item \textbf{Keep \texttt{p} consistent with the active ledger.}
    Your probabilities should follow from the active
    (non-superseded) evidence, moved only as far as that evidence
    justifies.
\end{itemize}

\PromptHeading{Output Format}

Conclude with two things, in order, each in its own tag:

\begin{enumerate}
    \item Your belief notebook, as a JSON object inside
    \texttt{<belief\_notebook>...</belief\_notebook>}.

    \item Your forecast, as a strict-JSON dictionary inside
    \texttt{<answer>...</answer>}---keys in double quotes, values
    numeric, no trailing commas. This is the only format the parser
    accepts, and its probabilities must equal your notebook's
    \texttt{p} exactly.
\end{enumerate}

\noindent\begin{minipage}{\linewidth}
\textbf{Example output (illustrative):}

\begin{PromptCode}
<belief_notebook>
{"assessment": {"p": {"YES": 0.32, "NO": 0.68},
"open_questions": ["Has the agency confirmed a revised timeline?"]},
"evidence_ledger": [
{"claim": "Regulator opened a formal review on 2025-05-12",
"supports": ["YES"],
"date_observed": "2025-05-13",
"status": "active",
"note": "official press release; primary source"},
{"claim": "Agency spokesperson said no decision is expected this quarter",
"rules_out": ["YES"],
"date_observed": "2025-05-20",
"status": "active",
"note": "direct quote, official"},
{"claim": "Early rumor of imminent approval",
"supports": ["YES"],
"date_observed": "2025-04-02",
"status": "superseded",
"note": "single blog; contradicted by the 05-12 review"}]}
</belief_notebook>
\end{PromptCode}
\end{minipage}\par

\noindent\begin{minipage}{\linewidth}
\textbf{Binary.}
Keys are exactly \texttt{"YES"} and \texttt{"NO"}, and values sum to
\(1\). Format example (numbers are illustrative):

\begin{PromptCode}
<answer>{"YES": 0.63, "NO": 0.37}</answer>
\end{PromptCode}
\end{minipage}\par

\noindent\begin{minipage}{\linewidth}
\textbf{Multiple-choice.}
Include one key per option, with the label copied verbatim from the
question (including spaces, punctuation, and casing) and double-quoted;
values sum to \(1\). Format example (labels and numbers are
illustrative):

\begin{PromptCode}
<answer>{"Manchester City FC": 0.33,
"Draw (Leeds United FC vs. Manchester City FC)": 0.17,
"Leeds United FC": 0.50}</answer>
\end{PromptCode}
\end{minipage}\par
\end{tcolorbox}
\nopagebreak[4]

\captionof{figure}{
Memory-on forecasting-agent system prompt.
}
\label{fig:memory_on_prompt}

\Needspace{12\baselineskip}
\subsection{Corpus-Side Event Filtering}
\label{sec:appendix:prompts:filtering}

\begin{tcolorbox}[system prompt={Query Decomposition}]
You are an expert forecasting analyst. Given an event forecasting
question, your job is to break it down into the specific information a
forecaster would need to assess it \textbf{before the question
resolves}.

\PromptHeading{Critical Constraint}

The agent that uses your queries can only see articles published
\textbf{before the forecast date} (i.e., before the question resolves).
Do \textbf{NOT} generate queries aimed at the \emph{outcome} of the
event---post-event reports, official resolution announcements, election
results, the actual launch price, the actual rate decision, etc. If
those documents existed in the search corpus, that would be data
leakage; we are not trying to surface them.

Generate queries aimed at material that legitimately exists
\emph{before} resolution: leading signals, expert analysis, prior
trends, structural context, and base rates.

\PromptHeading{What You Produce}

Return a JSON object with exactly two lists of search queries:

\begin{itemize}
    \item \texttt{evidence}: pre-resolution signals that move a
    forecaster's belief. Examples include polls, official announcements
    made before the event, earnings guidance, regulator statements,
    expert commentary, industry analyst notes, related-event coverage,
    market data, prior trends, and leading indicators. Time-anchor these
    queries where useful.

    \item \texttt{context}: time-independent background needed to
    interpret the question. Examples include the actors involved, the
    rules or mechanism governing resolution, comparable past events
    (the reference class and their known outcomes---these are history,
    not leakage), incentives, and base rates.
\end{itemize}

Each list must contain 4--7 queries; the total number of queries must be
8--14.

\PromptHeading{Rules for the Queries}

\begin{itemize}
    \item Use proper nouns wherever possible (entity names, locations,
    institution names). Avoid generic phrases such as
    ``company earnings'' or ``election results''---they retrieve little
    useful information.

    \item For evidence queries, date-anchor when the question is
    time-sensitive (e.g.,
    ``Apple iPhone 17 Pro pricing rumor August 2025'', rather than
    simply ``iPhone pricing'').

    \item Aim for diverse phrasings across the two lists. Do not repeat
    the same query in both angles. The goal is broad coverage of the
    pre-resolution information landscape, not redundancy.

    \item Each query should contain 3--10 words. Long queries match
    worse on dense retrieval and BM25 alike.

    \item Do \textbf{NOT} include the question's title verbatim. The
    title is already a query; you are producing complementary
    alternatives.

    \item Do \textbf{NOT} generate queries that aim at the
    post-resolution answer (e.g.,
    ``FOMC September 2025 rate decision announcement'' or
    ``iPhone 17 Pro launch price''). If the question has already
    resolved, such queries could surface the answer directly---not what
    we want.
\end{itemize}

\PromptHeading{Output Format}

Return \textbf{ONLY} a JSON object. No prose, no Markdown fences, and no
preamble.

\noindent\begin{minipage}{\linewidth}
\noindent\begin{minipage}{\linewidth}
Schema:

\begin{PromptCode}
{
  "evidence": ["q1", "q2", ...],
  "context":  ["q1", "q2", ...]
}
\end{PromptCode}
\end{minipage}\par
\end{minipage}\par

Each list must contain 4--7 strings. The total number of queries across
both lists should be 8--14.

\PromptHeading{Example}

\textbf{Question:} Will the Federal Reserve cut interest rates in
September 2025?

\textbf{Resolution criteria:} This market resolves YES if the FOMC
announces a rate cut at its September 2025 meeting. NO otherwise.

\noindent\begin{minipage}{\linewidth}
\textbf{Output:}

\begin{PromptCode}
{
  "evidence": [
    "Federal Reserve August 2025 CPI inflation report",
    "FOMC July 2025 minutes dot plot",
    "Jerome Powell Jackson Hole 2025 speech",
    "treasury yield curve August 2025",
    "Fed officials hawkish dovish remarks August 2025",
    "labor market jobs report August 2025"
  ],
  "context": [
    "FOMC voting members 2025 composition",
    "Federal Reserve dual mandate inflation employment",
    "Fed rate cut history 2024 pause cycle",
    "FOMC meeting calendar 2025 schedule"
  ]
}
\end{PromptCode}
\end{minipage}\par
\end{tcolorbox}
\nopagebreak[4]

\captionof{figure}{
Query-decomposition prompt used for corpus-side evidence retrieval.
}
\label{fig:query_decomposition_prompt}

\Needspace{12\baselineskip}
\begin{tcolorbox}[system prompt={Evidence-Sufficiency and Leakage Judge}]
You are an expert judge evaluating whether a small offline news corpus
carries enough signal to forecast an event question. You will be shown:

\begin{enumerate}
    \item The forecasting question (title, resolution criteria, and the
    recorded close date).

    \item A retrieval window
    \texttt{[corpus\_start\_date, close\_date - 2 days]}---the lower
    bound is the corpus's earliest indexed date; the upper bound leaves
    a two-day buffer before the recorded close date.

    \item A ranked list of articles retrieved from the offline corpus for
    this question.
\end{enumerate}

Your job is \textbf{not} to forecast the answer. Your job is to decide
whether a competent forecaster, \emph{given only these articles and
reasoning from them}, could form a calibrated belief about the question's
outcome---without relying on knowledge from outside the corpus.

\PromptHeading{What You Must Produce}

\noindent\begin{minipage}{\linewidth}
Return a single JSON object with this exact shape. No fences and no prose
around it:

\begin{PromptCode}
{
  "verdict": "SUFFICIENT" | "PARTIAL" | "INSUFFICIENT",
  "informativeness": 0 | 1 | 2 | 3,
  "leakage_detected": true | false,
  "leakage_reason": null | "<short string>",
  "confident_lean": null | "YES" | "NO" | "<leg-label>",
  "key_hits": ["<article_id>", ...],
  "reasoning": "<one or two sentences>"
}
\end{PromptCode}
\end{minipage}\par

\PromptHeading{How to Decide Each Field}

\PromptSubheading{\texttt{verdict} --- three buckets}

\begin{itemize}
    \item \texttt{SUFFICIENT} --- the corpus contains direct or strong
    indirect evidence about the question's outcome. A well-reasoned
    agent reading these articles would end up with a confident,
    well-calibrated belief (closer to \(0\) or \(1\), in the binary case)
    by resolution date.

    \item \texttt{PARTIAL} --- the corpus contains some signal:
    relevant background, related entities, partial evidence, or one weak
    indicator. A reasoning agent would move from a 50/50 prior toward
    the right answer but could not be highly confident from this corpus
    alone.

    \item \texttt{INSUFFICIENT} --- the corpus contains no useful
    signal: hits are off-topic, only superficially related, or the window
    is empty of substantive coverage. The agent would have no basis to
    update from the prior.
\end{itemize}

\PromptSubheading{\texttt{informativeness} --- 0--3 ordinal}

A finer-grained score representing how much these articles would move a
competent forecaster from a 50/50 prior:

\begin{itemize}
    \item \(0\) = no movement (corpus is noise, off-topic, or irrelevant).
    \item \(1\) = small movement (some related context, no decisive evidence).
    \item \(2\) = moderate movement (multiple corroborating leading
    indicators, or one strong indirect signal).
    \item \(3\) = large movement (one or more articles that, while
    published before resolution, essentially answer the question).
\end{itemize}

Map this to \texttt{verdict} consistently:

\begin{itemize}
    \item \(0 \rightarrow\) \texttt{INSUFFICIENT}
    \item \(1 \rightarrow\) \texttt{PARTIAL}
    \item \(2\) or \(3 \rightarrow\) \texttt{SUFFICIENT}
\end{itemize}

\PromptSubheading{Recency --- does the evidence cover the right time period?}

Some questions require \emph{recent} evidence to be informative; others
do not. Apply this test before scoring:

\begin{enumerate}
    \item Determine the question's \textbf{predictive horizon} from the
    title and resolution criteria.

    \begin{itemize}
        \item Snapshot or count over a fixed window
        (e.g., ``\# tweets Feb 17--24'', ``weekly views'',
        ``ranking on Feb 5'', ``price on close'')
        \(\rightarrow\) predictive horizon of approximately days to
        two weeks.

        \item Trajectory needing fresh data points
        (e.g., ``will X reach \$N by date'', ``will polls move K points'')
        \(\rightarrow\) requires evidence within roughly the last month.

        \item One-shot event whose preconditions persist
        (e.g., ``will Apple announce X'', ``will Russia capture town Y'')
        \(\rightarrow\) older articles describing the actors or mechanism
        are acceptable.
    \end{itemize}

    \item Compare against the \textbf{median age of the retrieved hits}
    in the \texttt{retrieval stats} line:
    \texttt{median\_hit\_age\_days},
    \texttt{freshest\_hit\_age\_days}, and
    \texttt{hits\_within\_30d}. The retrieval cutoff is
    \texttt{close\_date - 2 days}.

    \item If the question requires recent evidence but the corpus has
    none in the relevant window (e.g.,
    \texttt{hits\_within\_30d == 0} for a snapshot question), the
    verdict is \textbf{\texttt{INSUFFICIENT}} regardless of how many
    topically related but stale articles exist. Background articles
    about the entity do \textbf{not} substitute for missing recent data.
\end{enumerate}

When in doubt, a snapshot/count question with no hits in the last
approximately two weeks is \texttt{INSUFFICIENT}; a one-shot event
question with strong older articles can still be \texttt{SUFFICIENT}.

\PromptSubheading{\texttt{leakage\_detected} --- corpus integrity check}

Before judging informativeness, verify that the corpus respects the
retrieval window. The question's \texttt{close\_date} is the market's
\textbf{recorded close date}. If \textbf{ANY} hit's
\texttt{published\_date} is
\textbf{after} \texttt{close\_date - 2 days}, set
\texttt{leakage\_detected: true} and populate
\texttt{leakage\_reason} with a short string identifying the offending
article. Do \textbf{NOT} use those hits when judging sufficiency---score
the corpus \emph{as if those hits were absent}.

If an article dated before close already states the resolved outcome
(e.g., an article dated \texttt{close\_date - 2 days} that reports the match
score because the metadata date is the prior day's late edition), treat
that as content leakage and flag it in the same way.

\PromptSubheading{\texttt{confident\_lean} --- optional directional read}

If the in-window corpus gives a clear directional answer:

\begin{itemize}
    \item For binary questions, set \texttt{confident\_lean} to
    \texttt{"YES"} or \texttt{"NO"}.

    \item For multi-leg questions, set it to the leg label that the
    evidence most strongly supports (e.g.,
    \texttt{"Manchester City FC"}).

    \item If the evidence is mixed, weak, or absent, set it to
    \texttt{null}.
\end{itemize}

This is a sanity-check field---it allows us to measure later whether
\texttt{SUFFICIENT} verdicts actually correlate with correct calibration.

\PromptSubheading{\texttt{key\_hits} --- citations}

Provide up to five \texttt{article\_id} values from the retrieved hits
that most influenced your verdict. Order them by importance, with the
most decisive first. If
\texttt{verdict == "INSUFFICIENT"}, this can be an empty list.

\PromptSubheading{\texttt{reasoning} --- one or two sentences}

Provide a short justification for the verdict. Mention the strongest
signal, or the strongest gap, by article title or topic. Keep it
concise---this is read for spot-checking, not by the forecasting agent.

\PromptHeading{Important Rules}

\begin{itemize}
    \item Use \textbf{ONLY} the retrieved articles. Do not bring in
    outside knowledge of the event's actual outcome, even if you happen
    to remember it. The point of this judge is to score the
    \emph{corpus}, not your prior.

    \item Be conservative on \texttt{SUFFICIENT}. If the only
    ``evidence'' is one article that mentions the entities involved
    without addressing the resolution criteria, that is
    \texttt{PARTIAL} at best.

    \item Recognize the difference between \emph{relevant background}
    (\texttt{PARTIAL}) and \emph{predictive evidence}
    (\texttt{SUFFICIENT}). Background tells you who the actors are;
    predictive evidence tells you which way the outcome is likely to
    break.

    \item When in doubt between two adjacent buckets, choose the lower
    one:
    \[
    \texttt{INSUFFICIENT}
    <
    \texttt{PARTIAL}
    <
    \texttt{SUFFICIENT}.
    \]
\end{itemize}

\PromptHeading{Output Format Reminder}

Return \textbf{ONLY} the JSON object. No Markdown fences, no preamble,
and no prose outside the JSON. The object must validate against the
schema above.
\end{tcolorbox}
\nopagebreak[4]

\captionof{figure}{
Evidence-sufficiency and leakage judge prompt used for corpus-side
filtering.
}
\label{fig:evidence_judge_prompt}

\endgroup

\stopcontents[sections]

\end{document}